%% file: manuscript.tex
\documentclass[twoside]{article}
\usepackage[preprint]{twocolumnconf}

\usepackage[round]{natbib}

\usepackage{amsmath,amssymb,amsthm,mathtools,bm}
\usepackage{xspace}
\usepackage{booktabs}
\usepackage{graphicx}
\usepackage{pgf}
\providecommand{\mathdefault}[1]{#1}
\providecommand{\mathregular}[1]{#1}
\usepackage{enumitem}
\usepackage{hyperref}
\usepackage{xcolor}
\usepackage{float}

\usepackage{algorithm}
\usepackage{algorithmic}

\newtheorem{theorem}{Theorem}
\newtheorem{proposition}[theorem]{Proposition}

\newtheorem{corollary}[theorem]{Corollary}
\theoremstyle{definition}
\newtheorem{definition}[theorem]{Definition}

\theoremstyle{remark}

\DeclareMathOperator{\vol}{vol}
\DeclareMathOperator{\tr}{tr}
\DeclareMathOperator{\Scal}{Scal}

\newcommand{\R}{\mathbb{R}}

\newcommand{\E}{\mathbb{E}}

\newcommand{\M}{\mathcal{M}}
\newcommand{\dd}{\,\mathrm{d}}

\newcommand{\ess}{\mathrm{ESS}}
\newcommand{\gess}{\mathrm{gESS}}
\newcommand{\ip}[2]{\left\langle #1,#2\right\rangle}

\newcommand{\mmd}{\mathrm{MMD}}

\begin{document}

%
\runningtitle{ }

%
\runningauthor{ }

\twocolumn[

\aistatstitle{Heat-Kernel Entropy Profiles and Geometric Effective Sample Size for Weighted Measures on Manifolds}

\aistatsauthor{ Kisung You \And Boram Cho}
\aistatsaddress{ Baruch College \And Yale University} ]

\begin{abstract}
Weighted empirical measures on compact manifolds appear in importance sampling, particle approximations, posterior summaries, quadrature, and representation learning. Ordinary effective sample size and related weight summaries ignore the geometry of the support. We introduce heat-kernel entropy profiles to measure nonuniformity after intrinsic diffusion at a range of scales. For order-two R\'enyi entropy, pairwise heat-kernel overlaps give an exact profile and a geometric effective sample size. This effective sample size discounts nearby or duplicate particles. It approaches ordinary effective sample size as overlaps between distinct particles vanish. On compact boundaryless manifolds, we establish profile monotonicity, gESS scale limits, deterministic-weight consistency, and a bounded-ratio result for self-normalized importance sampling. On spheres, the unlogged profile decomposes into spherical-harmonic energies. The first terms are squared mean-resultant and traceless-second-moment energies, which give vMF- and Bingham-type scalar summaries. Experiments identify antipodal, girdle, multimodal, and duplicate-particle structures that weight-only and first-moment summaries miss.
\end{abstract}

\section{INTRODUCTION}
Let $\M$ be a compact Riemannian manifold and suppose a probability
measure of interest is represented by weighted atoms,
\begin{equation}
  \widehat P_w=\sum_{i=1}^n w_i\delta_{x_i},
  \quad x_i\in\M,\; w_i\ge0,\; \sum_iw_i=1.
  \label{eq:weighted_measure}
\end{equation}
Such measures arise in quadrature, importance sampling, particle methods,
posterior summaries, ensembles, and normalized representations. Our goal is to summarize the diffuseness of \eqref{eq:weighted_measure} using both its weights and support geometry.

Many standard diagnostics are label-based. The
ordinary effective sample size (ESS), used in importance sampling and
sequential Monte Carlo, monitors weight degeneracy
\citep{kong_1994_SequentialImputationsBayesian, liu_1998_SequentialMonteCarlo, delmoral_2006_SequentialMonteCarlo}. It is defined as
\begin{equation*}
  \ess(w)=\frac{1}{\sum_iw_i^2}.  
\end{equation*}
This quantity is also the inverse-Simpson, or order-two Hill, effective number of the
weights \citep{hill_1973_DiversityEvennessUnifying}. It diagnoses label-level
weight concentration but is not a geometric property of the represented
measure. Splitting an atom into identical labels should not increase its
diffuseness. Likewise, equally weighted particles should contribute differently when their locations are close rather than far apart.

Geometric uncertainty also depends on scale. A weighted measure may resemble
many particles at fine resolution, a few clusters at medium resolution, and
a nearly uniform measure after sufficient smoothing. On spheres, common first-moment summaries include mean resultant length and fitted von Mises-Fisher (vMF) concentration.
These summaries can miss antipodal, girdle-like, and symmetric multimodal structure
\citep{fisher_1953_DispersionSphere,mardia_2000_DirectionalStatistics,
banerjee_2005_ClusteringUnitHypersphere}. Our examples focus on these cases, but the method applies to general manifolds.

We apply intrinsic heat flow to the weighted atoms and measure nonuniformity over diffusion time. For order-two R\'enyi entropy, a pairwise heat-kernel identity avoids manifold integration. Normalizing these overlaps gives a geometric effective sample size (gESS) for each scale. The unlogged profile is also a heat-kernel maximum mean discrepancy (MMD) from uniformity. We interpret the resulting quantities as effective occupied volume and effective particle number. Small- and large-time results justify these interpretations.

The theory covers compact,
connected, boundaryless manifolds with normalized volume. Exact
implementation requires heat-kernel evaluations or controlled
approximations. Our experiments use $S^2$, where the heat kernel is explicit.
The vMF and Bingham interpretations are specific to spheres and do not enter the general definition.

\paragraph{Contributions.}
The paper makes five contributions. First, we define heat-kernel R\'enyi-2 profiles for weighted empirical measures on compact manifolds. Pairwise heat-kernel overlaps give an exact computation. Second, gESS discounts nearby and duplicate atoms, while approaching ordinary ESS when distinct-atom overlaps vanish. Third, small-time analysis describes local heat-kernel volume, curvature, duplicate coalescence, and two-atom merging. Fourth, a spectral expansion yields a spherical specialization. Its first two nonconstant energies are squared mean-resultant and traceless-second-moment energies. Finally, we prove stability and weighted-consistency results, including a bounded-ratio case for self-normalized importance sampling. The appendix contains proofs and further experiments. Replication code is available at \url{https://github.com/kisungyou/HeatKernelEntropyProfiles}.

\section{RELATED WORK}\label{sec:related-work}

\paragraph{Weight-only and similarity-aware effective numbers.}
Classical ESS diagnostics measure weight degeneracy in importance sampling and sequential Monte Carlo. Discrepancy-based alternatives address the same problem \citep{martino_2017_EffectiveSampleSize}. Hill numbers and inverse-Simpson diversity have an order-two effective-number interpretation \citep{hill_1973_DiversityEvennessUnifying}. Leinster-Cobbold diversity adds a user-specified similarity matrix \citep{leinster_2012_MeasuringDiversityImportance}. Our gESS is an order-two similarity diversity based on normalized heat-kernel overlap. This choice gives merge invariance, scale limits, and a heat asymptotic interpretation.

\paragraph{Heat smoothing, entropy, and kernel discrepancies.}
Kernel density estimation on spheres and compact manifolds has a substantial literature \citep{hall_1987_KernelDensityEstimation, pelletier_2005_KernelDensityEstimation, lebrigant_2019_ApproximationDensitiesRiemannian}. Nearest-neighbor entropy estimators are also available for hyperspherical data \citep{li_2011_KNearestNeighborBased}. These methods usually choose one smoothing scale to estimate a density or entropy. Here, diffusion scale determines the resolution. For \(A_P(t)=\iint k_{2t}(x,y)\dd P(x)\dd P(y)\), the unlogged profile satisfies \(A_P(t)-1=\mmd_{k_{2t}}^2(P,\nu)\). This identity links the profile to kernel mean embeddings \citep{sriperumbudur_2010_HilbertSpaceEmbeddings,gretton_2012_KernelTwoSampleTest, muandet_2017_KernelMeanEmbedding}. It also links the profile to Sobolev tests of uniformity on compact manifolds \citep{gine_1975_InvariantTestsUniformity, garcia-portugues_2018_OverviewUniformityTests}. Our transformations give the MMD curve two scale-dependent interpretations. They are an effective occupied volume fraction and an effective number of distinguishable atoms.

\paragraph{Spherical directional summaries.}
On spheres, vMF and Bingham/Fisher-Bingham models provide
classical first- and second-order summaries of directional structure \citep{kent_1982_FisherBinghamDistributionSphere,mardia_2000_DirectionalStatistics}. The proposed profile does not replace these parametric models in their unimodal or axial settings. Their moment energies appear as the first two nonconstant spectral components. Higher harmonic degrees distinguish structures that first- and second-order summaries cannot separate.

\section{HEAT ENTROPY PROFILES}

Throughout, $\M$ is a compact, connected, $m$-dimensional Riemannian manifold
without boundary. Let $\nu=\vol/\vol(\M)$ be normalized Riemannian volume, so
$\nu(\M)=1$ and the uniform density with respect to $\nu$ is identically one.
For vectors, $\|\cdot\|$ is the Euclidean norm. We use $\|\cdot\|_F$ for matrices and $\|\cdot\|_q$ for the $L^q(\nu)$ norm.
Let $\Delta$ be the Laplacian, with $\partial_t u=\Delta u$ and nonnegative eigenvalues of $-\Delta$. The heat kernel with respect to $\nu$ is $k_t(x,y)$
\citep{rosenberg_1997_LaplacianRiemannianManifold, grigoryan_2009_HeatKernelAnalysis}:
\begin{align*}
  \int_\M k_t(x,y)\dd\nu(y)&=1,\\
  k_{t+s}(x,y)&=\int_\M k_t(x,z)k_s(z,y)\dd\nu(z).
\end{align*}

For a probability measure $P$ on $\M$, write $H_tP$ for the heat-smoothed law.
For every $t>0$, $H_tP$ has a smooth density $p_t$ with respect to $\nu$.
Define
\begin{align*}
  D_{2,P}(t)&=\log\int_\M p_t(x)^2\dd\nu(x),\\
  U_{2,P}(t)&=\exp\{-D_{2,P}(t)\}.
\end{align*}
The quantity $D_{2,P}(t)$ is the order-two R\'enyi divergence of $H_tP$ from $\nu$. Its inverse exponential, $U_{2,P}(t)$, is the effective occupied volume fraction.

For the weighted empirical measure \eqref{eq:weighted_measure}, the
heat-smoothed density is
\begin{equation}
  \widehat p_t(y)=(H_t\widehat P_w)(y)
  =\sum_{i=1}^n w_i k_t(y,x_i).
  \label{eq:pt_def}
\end{equation}
Evaluating the same functional at $\widehat P_w$ gives the empirical heat entropy profile:
\[
  \widehat D_2(t)=\log\int_\M \widehat p_t(y)^2\dd\nu(y),
  \quad
  \widehat U_2(t)=\exp\{-\widehat D_2(t)\}.
\]
Write
\[
  \widehat A_w(t)=\int_\M \widehat p_t(y)^2\dd\nu(y),
\]
so that $\widehat D_2(t)=\log \widehat A_w(t)$ and
$\widehat U_2(t)=\widehat A_w(t)^{-1}$.

The quantity $U_{2,P}(t)$ measures effective occupied volume. For atomic input, $U_{2,\widehat P_w}(t)\to0$ as $t\downarrow0$ because the smoothed density becomes singular. The gESS defined below removes the self-overlap of each heat kernel. It therefore measures distinguishable atoms rather than volume.

\begin{theorem}[Pairwise identity and monotonicity]
\label{thm:pairwise_monotone}
For every $t>0$,
\begin{equation}
  \widehat A_w(t)
  =
  \int_\M \widehat p_t(y)^2\dd\nu(y)
  =
  \sum_{i,j=1}^n w_iw_jk_{2t}(x_i,x_j).
  \label{eq:pairwise_identity}
\end{equation}
Consequently,
\[
  \widehat U_2(t)
  =
  \left[\sum_{i,j}w_iw_jk_{2t}(x_i,x_j)\right]^{-1}.
\]
Moreover, $t\mapsto \widehat D_2(t)$ is nonincreasing and
$t\mapsto \widehat U_2(t)$ is nondecreasing. Finally,
\[
  \lim_{t\to\infty}\widehat D_2(t)=0,
  \qquad
  \lim_{t\to\infty}\widehat U_2(t)=1.
\]
\end{theorem}

Thus, intrinsic diffusion monotonically removes nonuniformity. The rate at which $\widehat U_2(t)$ approaches one defines the uncertainty profile.

\section{GEOMETRIC EFFECTIVE SAMPLE SIZE}\label{sec:gess}

The entropy profile measures effective volume. We obtain an effective number of atoms by normalizing pairwise heat overlaps by each kernel's self-overlap.

\begin{definition}[Geometric ESS]
\label{def:gess}
For $t>0$, define
\[
  s_t(x,y)=\frac{k_{2t}(x,y)}{\sqrt{k_{2t}(x,x)k_{2t}(y,y)}}
\]
and
\[
  \gess_w(t)=\left[\sum_{i,j}w_iw_js_t(x_i,x_j)\right]^{-1}.
\]
\end{definition}

On homogeneous manifolds, $k_{2t}(x,x)$ is constant. Examples include spheres and compact Lie groups with invariant metrics. Then
\[
  \gess_w(t)=k_{2t}(x,x)\widehat U_2(t).
\]
Thus, gESS measures effective volume in units of one heat kernel's self-volume. It is the order-two Hill/Leinster-Cobbold diversity for
$S_t=(s_t(x_i,x_j))_{ij}$ \citep{hill_1973_DiversityEvennessUnifying, leinster_2012_MeasuringDiversityImportance}.

\begin{theorem}[Properties and limits of gESS]
\label{thm:gess}
For every $t>0$,
\[
  0<s_t(x,y)\le1,
  \qquad
  s_t(x,x)=1.
\]
Consequently,
\[
  1\le \gess_w(t)\le \frac{1}{\sum_iw_i^2}=\ess(w).
\]
If exact duplicate atoms are merged by summing their weights, $\gess_w(t)$ is
unchanged. More generally, if the distinct support locations are
$z_1,\ldots,z_r$ with total masses
\[
  \alpha_a=\sum_{i:x_i=z_a}w_i,
\]
then
\[
  \lim_{t\downarrow0}\gess_w(t)=\frac{1}{\sum_{a=1}^r\alpha_a^2}.
\]
In particular, if all $x_i$ are distinct, the limit equals $\ess(w)$. Finally,
\[
  \lim_{t\to\infty}\gess_w(t)=1.
\]
\end{theorem}

Theorem~\ref{thm:gess} adds support geometry to ordinary ESS. Separated particles count individually at fine scales and become indistinguishable at coarse scales. Computation requires pairwise heat-kernel Gram matrices on a fixed grid of times. The cost is comparable to a kernel discrepancy or MMD diagnostic. Unlike one discrepancy value, gESS gives an effective-number curve. Algorithm~\ref{alg:heat-gess} combines Theorem~\ref{thm:pairwise_monotone} with the normalization in Definition~\ref{def:gess}.

\begin{algorithm}[t]
\caption{Heat entropy profile and geometric effective sample size}
\label{alg:heat-gess}
\small
\begin{algorithmic}[1]
\REQUIRE Weighted atoms $(x_i,w_i)_{i=1}^n$ with $\sum_iw_i=1$; heat scales
$0<t_1<\cdots<t_K$; heat kernel $k_t$ on $\M$.
\ENSURE Profiles $\{(t_k,\widehat U_2(t_k),\gess_w(t_k))\}_{k=1}^K$.
\FOR{$k=1,\ldots,K$}
  \STATE Form $G^{(k)}\in\mathbb R^{n\times n}$ with
  $G^{(k)}_{ij}\gets k_{2t_k}(x_i,x_j)$.
  \STATE Set $\widehat U_2(t_k)\gets \{w^\top G^{(k)}w\}^{-1}$.
  \IF{$\M$ is homogeneous}
    \STATE Set $S^{(k)}\gets G^{(k)}/k_{2t_k}(x_0,x_0)$ for any $x_0\in\M$.
  \ELSE
    \STATE Set $S^{(k)}_{ij}\gets
    G^{(k)}_{ij}/\sqrt{G^{(k)}_{ii}G^{(k)}_{jj}}$ for all $i,j$.
  \ENDIF
  \STATE Set $\gess_w(t_k)\gets\{w^\top S^{(k)}w\}^{-1}$.
\ENDFOR
\end{algorithmic}
\end{algorithm}

\begin{figure}[ht]
\centering
\input{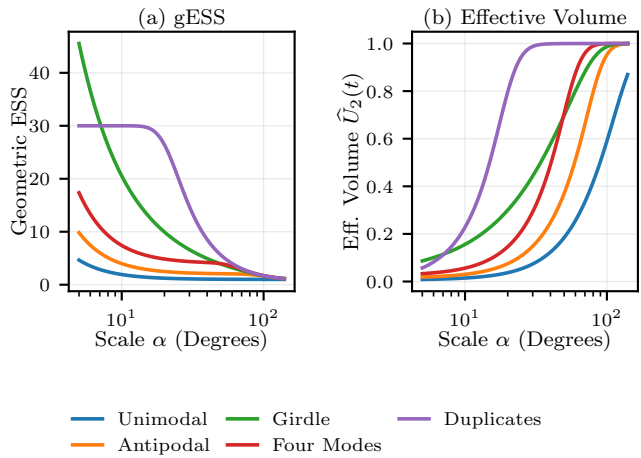}
\caption{Profiles for representative weighted measures on $S^2$. The horizontal axis is angular resolution $\alpha=\sqrt{8t}$ in degrees.
Panel (a) shows gESS. Panel (b) shows effective occupied volume.}
\label{fig:profiles}
\end{figure}

The two outputs have different interpretations. The $\widehat U_2$ profile is a normalized volume and measures effective occupancy after diffusion. The $\gess_w$ profile counts the heat-kernel footprints needed at a given scale. The same overlaps therefore give a density summary and a particle degeneracy diagnostic. Figure~\ref{fig:profiles} shows both quantities for the spherical examples.

\section{ASYMPTOTIC INTERPRETATION}

The next results describe the profiles at small and large diffusion scales. At small times, each atom resembles a heat ball. Nearby atoms then begin to overlap. At large times, heat flow leaves only low-frequency geometry. These limits provide the profile's scale calibration.

\subsection{Small-time behavior}

Let $V=\vol(\M)$. For simplicity, collapse exact duplicates in $\widehat P_w$ and write
\begin{equation*}
  \widehat P_w=\sum_{a=1}^r\alpha_a\delta_{z_a},
  \quad z_a\ne z_b\ (a\ne b),\; \sum_a\alpha_a=1,
\end{equation*}
and define the local mass quantities
\begin{equation*}
  S_\alpha=\sum_a\alpha_a^2,
  \quad
  C_\alpha=\sum_a\alpha_a^2 \Scal(z_a).
\end{equation*}
The normalized-volume heat kernel is $V$ times the Riemannian-volume kernel. For fixed distinct support points, off-diagonal terms decay faster than every power of $t$. The following fixed-configuration law is therefore determined by the diagonal heat expansion.

\begin{theorem}[Fixed-configuration small-time law]
\label{thm:small_time}
As $t\downarrow0$,
\begin{align}
  \widehat A_w(t)
  &:=\sum_{a,b}\alpha_a\alpha_b k_{2t}(z_a,z_b) \notag\\
  &=V(8\pi t)^{-m/2}
  \left[S_\alpha+\frac{t}{3}C_\alpha+O(t^2)+O(e^{-c/t})\right].
  \label{eq:small_time_expansion}
\end{align}
for some $c>0$ depending on the finite support. Consequently,
\begin{align*}
  \widehat D_2(t)
  &=\frac{m}{2}\log\frac{1}{8\pi t}+\log V+\log S_\alpha
    +\frac{t}{3}\frac{C_\alpha}{S_\alpha} \notag\\
  &\quad+O(t^2)+O(e^{-c/t}),\\
  \widehat U_2(t)
  &=\frac{(8\pi t)^{m/2}}{V}\frac{1}{S_\alpha}
    \left[1-\frac{t}{3}\frac{C_\alpha}{S_\alpha}+O(t^2)+O(e^{-c/t})\right].
\end{align*}
If all labels are distinct, then $\alpha_a=w_a$.
\end{theorem}

Theorem~\ref{thm:small_time} gives the asymptotic effective volume. The next proposition quantifies merging when pairwise distances are comparable to the diffusion radius.

\begin{proposition}[Local two-atom merging law]
\label{prop:merging}
Let $P=a\delta_x+(1-a)\delta_y$ with $0<a<1$ and $r=d_g(x,y)$. Fix $r_0<\operatorname{inj}(\M)$ and $C<\infty$. As $t\downarrow0$, the following expansion is uniform over $x,y$ satisfying $r\le r_0$ and $r^2\le Ct$:
\begin{equation}
  s_t(x,y)=\exp\left\{-\frac{r^2}{8t}\right\}\{1+O(t+r^2)\}.
  \label{eq:local_overlap}
\end{equation}
Hence
\begin{align}
  \gess_P(t)^{-1}
  &=a^2+(1-a)^2 \notag\\
  &\quad+2a(1-a)\exp\{-r^2/(8t)\}\notag\\
  &\qquad\times\{1+O(t+r^2)\}.
  \label{eq:two_atom_gess}
\end{align}
\end{proposition}

Atoms at distance $r$ merge when $r^2$ is of order $8t$. We therefore report the spherical scale as $\alpha=\sqrt{8t}$ radians. At $r=\alpha$, the leading overlap is $e^{-1}$.

\subsection{Large-time spectral behavior}

Let $0=\lambda_0<\lambda_1\le\lambda_2\le\cdots$ be the Laplace-Beltrami eigenvalues. Let $\{\varphi_r\}_{r\ge0}$ be an orthonormal eigenbasis in $L^2(\nu)$, with $\varphi_0\equiv1$. Define
\begin{equation*}
  \widehat a_r=\sum_iw_i\varphi_r(x_i),\qquad r\ge1.
\end{equation*}
The heat-kernel spectral representation gives the next result
\citep{rosenberg_1997_LaplacianRiemannianManifold, grigoryan_2009_HeatKernelAnalysis}.

\begin{theorem}[Spectral expansion]
\label{thm:spectral}
For every $t>0$,
\begin{equation}
  \widehat p_t-1=\sum_{r\ge1}e^{-\lambda_rt}\widehat a_r\varphi_r,
  \label{eq:pt_spectral}
\end{equation}
and
\begin{equation}
  \widehat D_2(t)=\log\left(1+\sum_{r\ge1}e^{-2\lambda_rt}\widehat a_r^2\right).
  \label{eq:D2_spectral}
\end{equation}
Equivalently,
\begin{align}
  \widehat A_w(t)-1
  &=\frac{1}{\widehat U_2(t)}-1 \notag\\
  &=\sum_{r\ge1}e^{-2\lambda_rt}\widehat a_r^2.
  \label{eq:Aminus_spectral}
\end{align}
Consequently, as $t\to\infty$, if $\lambda_+$ is the next distinct eigenvalue after $\lambda_1$, then
\begin{align*}
  \widehat D_2(t)
  &=e^{-2\lambda_1t}\sum_{\lambda_r=\lambda_1}\widehat a_r^2 \notag\\
  &\quad+O\left(e^{-2\lambda_+t}+e^{-4\lambda_1t}\right).
\end{align*}
\end{theorem}

The exact expansion \eqref{eq:Aminus_spectral} has no logarithmic cross terms. Its lowest nonzero eigenmodes control large-scale uncertainty and persist longest under diffusion.

\section{SPHERICAL SPECIALIZATION: vMF AND BINGHAM STRUCTURE}\label{sec:spheres}

We next specialize the construction to the sphere $S^{p-1}$, where $m=p-1$. Its first Laplace-Beltrami eigenspaces have standard interpretations in directional statistics. Degree one corresponds to a preferred direction. Degree two describes axial or elliptical departures from uniformity. Higher degrees describe multimodality and finer angular structure. Let $\nu$ be normalized surface measure. The eigenvalues are
\begin{equation*}
  \lambda_\ell=\ell(\ell+p-2),\qquad \ell=0,1,2,\ldots.
\end{equation*}
The eigenspaces consist of degree-$\ell$ spherical harmonics. Let $\widehat a_{\ell j}=\sum_iw_iY_{\ell j}(x_i)$ in an orthonormal basis of the degree-$\ell$ eigenspace. This is the double-index form of Theorem~\ref{thm:spectral}. A sum over $r$ with $\lambda_r=\lambda_\ell$ becomes $\sum_j\widehat a_{\ell j}^2$.

Let
\begin{equation*}
  \bar x_w=\sum_iw_ix_i,
  \quad
  Q_w=\sum_iw_ix_ix_i^\top.
\end{equation*}
The degree-1 spectral energy is
\begin{equation}
  B_1:=\sum_j\widehat a_{1j}^2=p\|\bar x_w\|^2,
  \label{eq:B1}
\end{equation}
and the degree-2 spectral energy is
\begin{equation}
  B_2:=\sum_j\widehat a_{2j}^2=\frac{p(p+2)}{2}\left\|Q_w-\frac{I_p}{p}\right\|_F^2.
  \label{eq:B2}
\end{equation}

\begin{corollary}[Spherical power expansion]
\label{cor:sphere}
On $S^{p-1}$,
\begin{align*}
  \widehat A_w(t)-1
  &=\frac{1}{\widehat U_2(t)}-1 \notag\\
  &=\sum_{\ell\ge1}e^{-2\ell(\ell+p-2)t}\sum_j\widehat a_{\ell j}^2 \notag\\
  &=e^{-2(p-1)t}B_1+e^{-4pt}B_2+
  \sum_{\ell\ge3}e^{-2\ell(\ell+p-2)t}B_\ell,
\end{align*}
where $B_\ell=\sum_j\widehat a_{\ell j}^2$.
Moreover,
\begin{equation*}
  \widehat D_2(t)=\log\{1+\widehat A_w(t)-1\}.
\end{equation*}
Thus, when $B_1\ne0$, the logged entropy contains the cross term
\begin{equation*}
  -\frac12e^{-4(p-1)t}B_1^2
\end{equation*}
before the displayed degree-2 term. If $B_1=0$, the $B_2$ term is the leading logged contribution whenever degree two is nonzero.
\end{corollary}

Up to $p$, $B_1$ is the squared weighted mean resultant length used by Rayleigh tests and vMF concentration fitting. The term $B_2$ measures traceless second-moment anisotropy associated with Bingham or Fisher-Bingham structure \citep{kent_1982_FisherBinghamDistributionSphere}. These energies retain neither mean direction nor anisotropy axes. Higher degrees measure multimodality and finer angular structure. Corollary~\ref{cor:sphere} identifies the first two coefficients of $\widehat A_w(t)-1$ as vMF- and Bingham-type energies. The logarithm introduces interactions among lower-degree terms.

\section{WEIGHTED CONSISTENCY}

So far, the profile has been a deterministic functional of the weighted measure. In applications, weighted atoms may approximate a target distribution. We establish uniform convergence on scale intervals bounded away from zero. This lower cutoff is needed because heat kernels become singular as $t$ approaches zero. Let
\begin{align*}
  A_P(t)&=\iint k_{2t}(x,y)\dd P(x)\dd P(y),  \\
  \widehat A_w(t)&=\sum_{i,j}w_iw_jk_{2t}(x_i,x_j),
\end{align*}
and
\begin{equation*}
    D_{2,P}(t)=\log A_P(t).
\end{equation*}

\begin{theorem}[Stability and weighted convergence]
\label{thm:consistency}
Fix $0<t_0<T<\infty$. There is a constant $C=C(t_0,T,\M)$ for which every pair of probability measures $P,Q$ satisfies
\begin{equation}
  \sup_{t\in[t_0,T]} |A_P(t)-A_Q(t)|
  \le C W_1(P,Q),
  \label{eq:W1_stability}
\end{equation}
where $W_1$ is the geodesic Wasserstein-1 distance. Consequently, if $\widehat P_w\to P$ in $W_1$, then
\begin{equation*}
  \sup_{t\in[t_0,T]} |\widehat D_2(t)-D_{2,P}(t)|\to0.
\end{equation*}
If $X_1,\ldots,X_n\overset{iid}{\sim}P$ and deterministic weights satisfy $S_n:=\sum_iw_i^2\to0$, then
\begin{equation}
  \sup_{t\in[t_0,T]} |\widehat A_w(t)-A_P(t)|
  =O_p\left(\sqrt{S_n\log(1/S_n)}\right),
  \label{eq:weighted_rate}
\end{equation}
and the same rate holds for $\widehat D_2(t)$.

Finally, let $X_i\overset{iid}{\sim}Q$. Suppose $P$ has density ratio $\rho=\dd P/\dd Q$, where $0\le\rho\le M$ and $\E_Q\rho=1$. When the denominator is positive, set $w_i=\rho(X_i)/\sum_j\rho(X_j)$. Use any fixed convention otherwise. Then
\begin{equation*}
  \sup_{t\in[t_0,T]} |\widehat A_w(t)-A_P(t)|=O_p\left(\sqrt{\frac{\log n}{n}}\right),
\end{equation*}
with the same rate for $\widehat D_2(t)$.
\end{theorem}

The deterministic-weight rate is a worst-case bound. If $P=\nu$, the linear Hoeffding projection vanishes. At each fixed $t$, the error then improves to $O_p(S_n)$. The constants deteriorate as $t_0\downarrow0$ because heat-kernel suprema and derivatives scale as powers of $t_0^{-1}$. Fine-scale resolution therefore has weaker statistical stability.

\section{EXPERIMENTS}\label{sec:experiments}

\begin{figure*}[ht]
\centering
\input{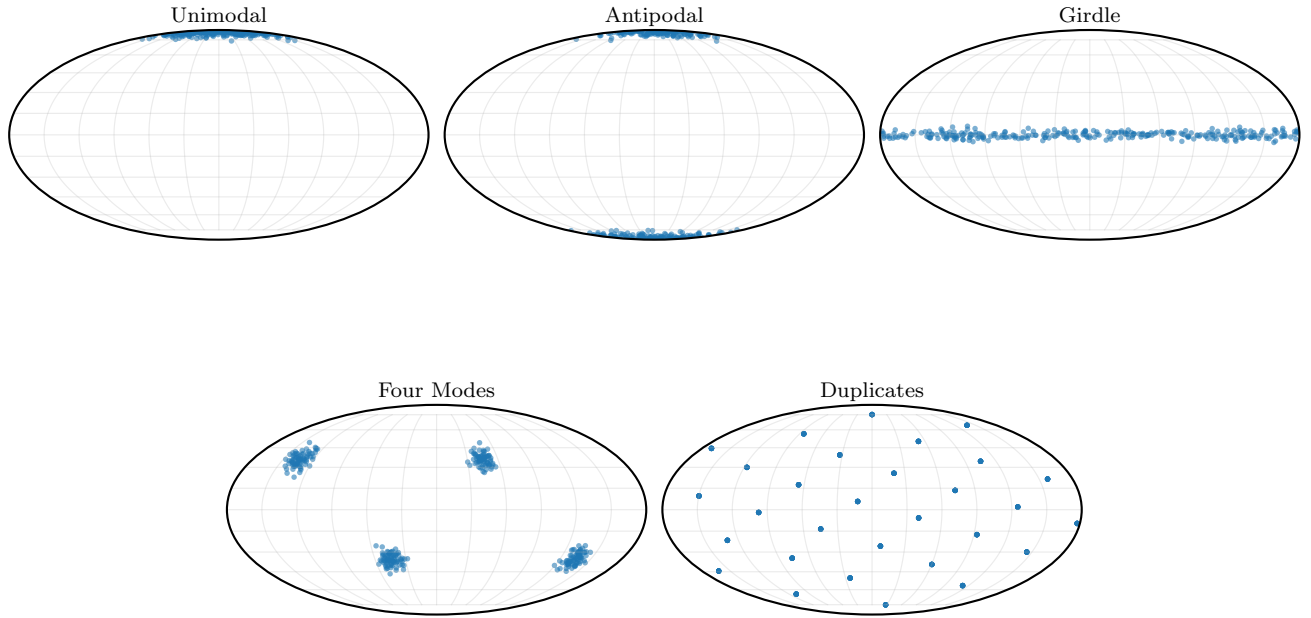}
\caption{Mollweide projections of the five synthetic $S^2$ configurations. Antipodal, girdle, and four-mode structures can have small resultant length. However, they remain highly nonuniform at relevant scales.}
\label{fig:clouds}
\end{figure*}

The theory applies to general manifolds, while the experiments use $S^2$ because its heat kernel is explicit. Spherical data also make first-moment failures easy to visualize. The synthetic configurations are chosen to expose limitations of weight-only ESS and first-moment summaries. Figure~\ref{fig:clouds} shows these configurations.

\begin{figure*}[t]
\centering
\input{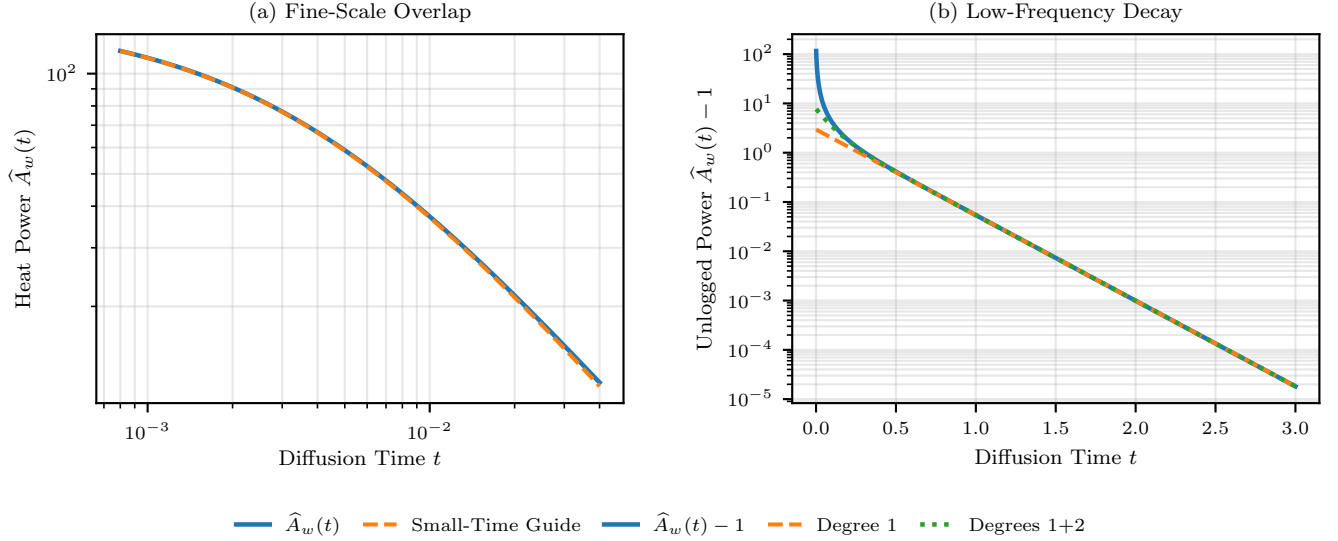}
\caption{Two-regime validation for a unimodal vMF cloud on $S^2$. Panel (a) compares $\widehat A_w(t)$ to the leading small-time pairwise heat-overlap approximation. Panel (b) compares $\widehat A_w(t)-1$ to the low-frequency approximation. The unlogged power is used because it is the cross-term-free object in the spherical expansion.}
\label{fig:asymptotics}
\end{figure*}

The synthetic values use the constructions in Appendix~\ref{app:add_experiments}. We compute each quantity with the routine in Section~\ref{sec:gess}. Only the heat Gram matrix is sphere-specific. The heat kernel is
\begin{equation*}
  k_t(x,y)=\sum_{\ell=0}^{\infty}(2\ell+1)e^{-\ell(\ell+1)t}P_\ell(x^\top y),
\end{equation*}
where $P_\ell$ is the Legendre polynomial. For each weighted cloud, we first compute
\begin{equation*}
  B_\ell=(2\ell+1)\sum_{i,j}w_iw_jP_\ell(x_i^\top x_j)
\end{equation*}
once and evaluate
\begin{equation*}
  \widehat A_w(t)=\sum_{\ell\ge0}e^{-2\ell(\ell+1)t}B_\ell.
\end{equation*}
As a similarity-sensitive baseline, we use Leinster-Cobbold-style diversity with geodesic Gaussian similarity:
\begin{equation*}
  D_{\rm LC}^{(\alpha)}=\left[\sum_{i,j}w_iw_j\exp\{-(d_g(x_i,x_j)/\alpha)^2\}\right]^{-1}.
\end{equation*}
This baseline matches the $e^{-1}$ convention at $d_g=\alpha$, but it is not derived from the heat semigroup.
We also report ordinary $\ell_2$ ESS. The real-data table includes entropy/perplexity ESS, $\ess_H=\exp\{-\sum_iw_i\log w_i\}$. This weight-only baseline appears in ESS surveys \citep{martino_2017_EffectiveSampleSize}. The heat-kernel MMD curve equals $\widehat A_w(t)-1$, as noted in Section~\ref{sec:related-work}. Figure~\ref{fig:asymptotics} uses this unlogged curve to assess both asymptotic approximations.

\paragraph{Synthetic configurations.}

Each synthetic example targets a different limitation. The unimodal cloud is a concentrated reference. The antipodal and girdle examples have nearly zero first moments but remain nonuniform. The four-mode example tests sensitivity to modal separation. The duplicate example tests atom-splitting invariance, which ordinary ESS lacks. We compare five equal-weight configurations with $n=300$. These are a vMF cloud, an antipodal mixture, a girdle, four separated modes, and duplicates at 30 quasi-uniform locations. Table~\ref{tab:synthetic} reports selected scales. Ordinary ESS is 300 in every row, while $R$ misses symmetric nonuniformity. The energy $B_2$ detects axial structure in the antipodal and girdle examples. In contrast, gESS gives a scale-dependent effective number. The duplicate example has 300 labels at only 30 locations. At $15^\circ$, its gESS is about 30, consistent with merge invariance and limited overlap between locations.

\begin{table}[t]
\centering
\caption{Synthetic examples on $S^2$. Here $R=\|\sum_iw_ix_i\|$ and $B_2$ is the degree-2/Bingham energy.
The baseline is $D_{\rm LC}^{(15)}$, and $\alpha=\sqrt{8t}$ is the angular scale in degrees.}
\label{tab:synthetic}
\small
\resizebox{\columnwidth}{!}{%
\begin{tabular}{lrrrrrr}
\toprule
Example & $R$ & $B_2$ & ESS & $\gess(15^\circ)$ & $D_{\rm LC}^{(15)}$ & $\gess(45^\circ)$ \\
\midrule
Unimodal vMF & 0.993 & 4.79 & 300 & 1.4 & 1.4 & 1.0 \\
Antipodal & 0.006 & 4.78 & 300 & 2.9 & 2.9 & 2.1 \\
Girdle & 0.029 & 1.24 & 300 & 13.3 & 13.4 & 4.4 \\
Four modes & 0.007 & 0.00 & 300 & 5.5 & 5.5 & 4.1 \\
Duplicates & 0.005 & 0.00 & 300 & 29.7 & 29.7 & 6.8 \\
\bottomrule
\end{tabular}%
}
\end{table}

\paragraph{Real-data embedding illustration.}

For a data example, we use the handwritten digits dataset\footnote{\url{https://doi.org/10.24432/C50P49}} from
\textsf{scikit-learn} \citep{pedregosa_2011_ScikitlearnMachineLearning}. The $8\times8$ images are standardized and projected onto three principal components. We normalize these projections onto $S^2$. The example compares reweighting the same embeddings with restricting them to a subset. Table~\ref{tab:digits} shows that nonuniform reliability-style weights reduce ordinary and geometric ESS. The 0-vs-1 subset also differs from the full cloud in directional structure.

\begin{table}[t]
\centering
\caption{Normalized digit embeddings after PCA to $\R^3$ and projection onto $S^2$. The weighted row uses deterministic weights proportional to $\exp(2.5x_1)$.}
\label{tab:digits}
\small
\resizebox{\columnwidth}{!}{%
\begin{tabular}{lrrrrrrr}
\toprule
Data & $n$ & $R$ & $B_2$ & ESS & $\ess_H$ & $\gess(15^\circ)$ & $D_{\rm LC}^{(15)}$ \\
\midrule
Digits all & 300 & 0.152 & 0.07 & 300 & 300 & 36.7 & 36.8 \\
Digits weighted & 300 & 0.649 & 0.53 & 123 & 156 & 15.4 & 15.5 \\
Digits 0/1 & 60 & 0.540 & 0.81 & 60 & 60 & 9.8 & 9.9 \\
\bottomrule
\end{tabular}%
}
\end{table}

\section{DISCUSSION}

Heat-kernel entropy profiles summarize weighted empirical measures on compact manifolds. For order-two R\'enyi entropy, the pairwise formula also gives a geometry-aware ESS. Its scale limits do not require a spherical parametrization. The fine-scale gESS limit is the ESS of coalesced support locations. At large times, the expansion identifies structures that persist under diffusion.

The theory assumes a compact, connected, boundaryless manifold with normalized Riemannian volume. Theorem~\ref{thm:pairwise_monotone} proves monotonicity for $\widehat U_2(t)$. It does not prove gESS monotonicity on nonhomogeneous manifolds because the diagonal normalization varies with location. Exact computation also requires heat-kernel evaluations. Harmonic expansions are natural on spheres and compact Lie groups. General manifolds may instead use spectral truncation, graph Laplacians \citep{chung_1997_SpectralGraphTheory}, diffusion maps \citep{coifman_2006_DiffusionMaps}, or numerical heat solvers.

The statistical results require a positive lower scale. Small scales reveal fine structure but have weaker stability because heat-kernel suprema and derivatives grow as $t\downarrow0$. We therefore state consistency on $[t_0,T]$ with $t_0>0$. The experiments describe profile behavior without calibrated inference or confidence bands. Further work could develop full-profile inference and allow heavy-tailed importance weights. Extensions to noncompact manifolds would also require a suitable reference measure. Another direction is joint use with maximum-entropy hierarchies for model-based uncertainty decomposition.



\bibliographystyle{apalike}
\bibliography{references}


\appendix


\section{PROOFS}\label{app:proofs}

\subsection{Proof of Theorem~\ref{thm:pairwise_monotone}}
By bilinearity and the definition of $\widehat p_t$,
\begin{equation*}
\int \widehat p_t(y)^2\dd\nu(y)
=\sum_{i,j}w_iw_j\int k_t(y,x_i)k_t(y,x_j)\dd\nu(y).
\end{equation*}
The heat kernel is symmetric and satisfies the semigroup identity. Hence, the integral is $k_{2t}(x_i,x_j)$, proving \eqref{eq:pairwise_identity}.

For monotonicity, $\widehat p_t$ solves $\partial_t\widehat p_t=\Delta\widehat p_t$. Under our convention, the spectrum of $-\Delta$ is nonnegative. Since $\M$ is compact without boundary,
\begin{align*}
  \frac{d}{dt}\int \widehat p_t^2\dd\nu
  &=2\int \widehat p_t\Delta\widehat p_t\dd\nu
   =-2\int\|\nabla\widehat p_t\|^2\dd\nu\le0.
\end{align*}
Therefore, $\widehat D_2(t)=\log\|\widehat p_t\|_2^2$ is nonincreasing, while $\widehat U_2(t)$ is nondecreasing. Compactness and connectedness imply that heat flow converges in $L^2(\nu)$ to the constant density $1$. Thus, $\|\widehat p_t\|_2^2\to1$.

\subsection{Proof of Theorem~\ref{thm:gess}}
The semigroup identity gives
\begin{align*}
  k_{2t}(x,y)&=\int k_t(x,z)k_t(y,z)\dd\nu(z)\\
  &=\ip{k_t(x,\cdot)}{k_t(y,\cdot)}_{L^2(\nu)}.
\end{align*}
The heat kernel is strictly positive on a connected compact manifold for $t>0$. Hence, $s_t(x,y)>0$. The Cauchy-Schwarz inequality gives
\begin{equation*}
  k_{2t}(x,y)^2\le k_{2t}(x,x)k_{2t}(y,y),
\end{equation*}
which proves $s_t(x,y)\le1$ and $s_t(x,x)=1$.

Since $s_t(x_i,x_j)\le1$ and $\sum_{i,j}w_iw_j=1$,
\begin{equation*}
  \sum_{i,j}w_iw_js_t(x_i,x_j)\le1,
\end{equation*}
so $\gess_w(t)\ge1$. Since $s_t(x_i,x_i)=1$ and off-diagonal terms are nonnegative,
\begin{equation*}
  \sum_{i,j}w_iw_js_t(x_i,x_j)\ge\sum_iw_i^2,
\end{equation*}
so $\gess_w(t)\le1/\sum_iw_i^2$.

Merging duplicate atoms leaves the double sum unchanged because their terms have the same similarity value. For distinct support points $z_a$ with masses $\alpha_a$, the denominator is
\begin{equation*}
  \sum_{a,b}\alpha_a\alpha_bs_t(z_a,z_b).
\end{equation*}
As $t\downarrow0$, $s_t(z_a,z_a)=1$, while $s_t(z_a,z_b)\to0$ for $a\ne b$. Gaussian upper and diagonal lower bounds make this convergence uniform away from the diagonal \citep{grigoryan_2009_HeatKernelAnalysis}. The denominator therefore converges to $\sum_a\alpha_a^2$. As $t\to\infty$, the spectral expansion and gap give uniform convergence of $k_{2t}(x,y)$ to $1$. Hence, $s_t(x,y)\to1$, and the denominator converges to $1$.

\subsection{Proof of Theorem~\ref{thm:small_time} and Proposition~\ref{prop:merging}}
Let $K_t$ be the heat kernel with respect to Riemannian volume. Since $\nu=\vol/V$, the normalized-volume kernel is $k_t=VK_t$. The diagonal Minakshisundaram-Pleijel expansion gives
\citep{rosenberg_1997_LaplacianRiemannianManifold, grigoryan_2009_HeatKernelAnalysis}
\begin{equation*}
  K_s(x,x)=(4\pi s)^{-m/2}\left[1+\frac{s}{6}\Scal(x)+O(s^2)\right].
\end{equation*}
With $s=2t$,
\begin{equation*}
  k_{2t}(z_a,z_a)=V(8\pi t)^{-m/2}\left[1+\frac{t}{3}\Scal(z_a)+O(t^2)\right].
\end{equation*}
The finite, distinct support points $z_a$ have a positive minimum distance. Gaussian heat-kernel bounds make the off-diagonal contribution $V(8\pi t)^{-m/2}O(e^{-c/t})$ for some $c>0$ \citep{grigoryan_2009_HeatKernelAnalysis}. Summing the diagonal terms gives \eqref{eq:small_time_expansion}. Taylor expansion of $\log(S_\alpha+\epsilon)$ and $(S_\alpha+\epsilon)^{-1}$ gives the remaining statements. Here, $\epsilon=(t/3)C_\alpha+O(t^2)+O(e^{-c/t})$.

For Proposition~\ref{prop:merging}, use the near-diagonal heat-kernel parametrix uniformly in a normal neighborhood:
\begin{equation*}
  K_s(x,y)=(4\pi s)^{-m/2}\exp\left\{-\frac{d_g(x,y)^2}{4s}\right\} \{u_0(x,y)+O(s)\}.
\end{equation*}
Here, $u_0(x,y)$ is the square root of the inverse Jacobian determinant of the exponential map. It satisfies $u_0(x,y)=1+O(d_g(x,y)^2)$. Combine this relation with the diagonal expansion in the denominator of $s_t(x,y)$, and set $s=2t$. This gives \eqref{eq:local_overlap}. Substitution into $a^2+(1-a)^2+2a(1-a)s_t(x,y)$ proves \eqref{eq:two_atom_gess}.

\subsection{Proof of Theorem~\ref{thm:spectral}}
The heat kernel has the spectral representation
\begin{equation*}
  k_t(x,y)=\sum_{r\ge0}e^{-\lambda_rt}\varphi_r(x)\varphi_r(y),
\end{equation*}
where $\varphi_0\equiv1$. Substituting this into \eqref{eq:pt_def} gives
\begin{equation*}
  \widehat p_t(y)=1+\sum_{r\ge1}e^{-\lambda_rt}\left(\sum_iw_i\varphi_r(x_i)\right)\varphi_r(y),
\end{equation*}
which is \eqref{eq:pt_spectral}. Orthogonality implies
\begin{equation*}
  \int\widehat p_t^2\dd\nu=1+\sum_{r\ge1}e^{-2\lambda_rt}\widehat a_r^2,
\end{equation*}
which proves \eqref{eq:D2_spectral} and \eqref{eq:Aminus_spectral}. Applying $\log(1+z)=z+O(z^2)$ and separating the lowest nonzero eigenspace gives the large-time expansion.

\subsection{Proof of Corollary~\ref{cor:sphere}}
\label{app:sphere}
For $S^{p-1}$ with normalized surface measure, the coordinate functions satisfy
\begin{equation*}
  \int x_ax_b\dd\nu(x)=\frac{\delta_{ab}}{p}.
\end{equation*}
Thus, $\sqrt p\,x_a$ for $a=1,\ldots,p$ form an orthonormal basis of the degree-1 eigenspace. Its energy is
\begin{equation*}
  \sum_{a=1}^p\left(\sum_iw_i\sqrt p\,x_{ia}\right)^2=p\left\|\sum_iw_ix_i\right\|^2.
\end{equation*}

For degree 2, let $A$ and $B$ be symmetric traceless matrices. The fourth-moment identity on the sphere gives
\begin{equation*}
  \int (x^\top A x)(x^\top Bx)\dd\nu(x)
  =\frac{2\tr(AB)}{p(p+2)}.
\end{equation*}
Hence, $A\mapsto \sqrt{p(p+2)/2}\,x^\top A x$ is an isometry into the degree-2 harmonic subspace. Its domain is the Frobenius space of symmetric traceless matrices. The map is onto because both spaces have dimension $p(p+1)/2-1$. Since $Q_w-I_p/p$ is the traceless part of $Q_w$, the degree-2 energy is
\begin{equation*}
  \frac{p(p+2)}{2}\left\|Q_w-\frac{I_p}{p}\right\|_F^2.
\end{equation*}
This proves \eqref{eq:B1} and \eqref{eq:B2}.

\subsection{Proof of Theorem~\ref{thm:consistency}}
\label{app:weighted_rate}
We first prove deterministic stability. For fixed $t\in[t_0,T]$, define
\begin{equation*}
  f_{P,t}(x)=\int k_{2t}(x,y)\dd P(y).
\end{equation*}
Compactness and $t\ge t_0>0$ uniformly bound $k_{2t}$ and its first spatial derivatives on $\M^2\times[t_0,T]$. Thus, $f_{P,t}$ is uniformly Lipschitz with constant at most $L=L(t_0,T,\M)$. For probability measures $P,Q$,
\begin{align*}
  A_P(t)-A_Q(t)
  &=\iint k_{2t}(x,y)\dd(P-Q)(x)\dd P(y) \\
  &\quad+\iint k_{2t}(x,y)\dd Q(x)\dd(P-Q)(y).
\end{align*}
Kantorovich-Rubinstein duality bounds each term by $L W_1(P,Q)$. This proves \eqref{eq:W1_stability} with $C=2L$, uniformly over $[t_0,T]$. Since $A_P(t)=\|p_t\|_2^2\ge1$, the logarithm is one-Lipschitz on the relevant range.

For the random weighted rate, set $g_t(x,y)=k_{2t}(x,y)$, write
$S_n=\sum_iw_i^2$, and let $F_n(t)=\widehat A_w(t)$. Smoothness on the
compact set $\M^2\times[t_0,T]$ gives $0<g_t\le B$ and a uniform
time-Lipschitz constant $L_t$. If
$d_P(t)=\E\{g_t(X,X)\}$, independence gives
\begin{equation*}
  \E F_n(t)=(1-S_n)A_P(t)+S_nd_P(t).
\end{equation*}
Hence $\sup_t|\E F_n(t)-A_P(t)|\le C S_n$.

Replacing $X_i$ while holding the other observations fixed changes
$F_n(t)$ by at most $4Bw_i$. McDiarmid's inequality therefore yields,
uniformly at every fixed $t$,
\begin{equation}
  \Pr\{|F_n(t)-\E F_n(t)|>u\}
  \le 2\exp\{-c u^2/(B^2S_n)\}.
  \label{eq:mcdiarmid_weighted}
\end{equation}
Choose a time grid of mesh $u/(4L_t)$ and cardinality at most $C/u$.
Both $F_n(t)$ and $\E F_n(t)$ are uniformly Lipschitz in $t$. Taking
$u=K\sqrt{S_n\log(1/S_n)}$ in \eqref{eq:mcdiarmid_weighted}, applying a
union bound on the grid, and then interpolating proves
\begin{equation*}
  \sup_{t\in[t_0,T]}|F_n(t)-\E F_n(t)|
  =O_p\{\sqrt{S_n\log(1/S_n)}\}.
\end{equation*}
Since $S_n=o\{\sqrt{S_n\log(1/S_n)}\}$, the bias bound proves
\eqref{eq:weighted_rate}.

Now suppose $P=\nu$ and fix $t>0$. Then $A_\nu(t)=1$ and
$\int g_t(x,y)\dd\nu(y)=1$, so $g_t(x,y)-1$ is canonical. The
off-diagonal part of $F_n(t)-1$ has variance at most
$C\sum_{i\ne j}w_i^2w_j^2\le CS_n^2$. The diagonal part has mean
$O(S_n)$ and variance at most $C\sum_iw_i^4\le CS_n^2$. Thus
$F_n(t)-1=O_p(S_n)$, as asserted in the main text.

For self-normalized importance sampling, write $Z_n=n^{-1}\sum_i\rho(X_i)$ and
\begin{equation*}
  N_n(t)=n^{-2}\sum_{i,j}\rho(X_i)\rho(X_j)g_t(X_i,X_j).
\end{equation*}
The off-diagonal expectation of $N_n(t)$ is $(1-n^{-1})A_P(t)$, and
the bounded diagonal contributes $O(n^{-1})$ uniformly in $t$. Replacing
one observation changes $N_n(t)$ by at most $C/n$. The same grid and
bounded-differences argument therefore gives
\begin{equation*}
  \sup_t|N_n(t)-A_P(t)|=O_p\{\sqrt{\log n/n}\}.
\end{equation*}
Boundedness of $\rho$ also gives $Z_n=1+O_p(n^{-1/2})$. On $Z_n>0$, $\widehat A_w(t)=N_n(t)/Z_n^2$. Uniform boundedness of $A_P(t)$ gives the claimed rate for this ratio. Finally, $Q\{\rho(X)=0\}<1$, so $\Pr(Z_n=0)=Q\{\rho(X)=0\}^n\to0$. The convention on this event is asymptotically irrelevant.

\section{ADDITIONAL EXPERIMENTS AND REPRODUCIBILITY}
\label{app:add_experiments}

This appendix gives the details needed to reproduce the figures and tables. It also reports results beyond the main-text summaries. Replications assess profile stability, while a truncation study measures numerical sensitivity. A self-normalized importance-sampling example compares ordinary ESS with gESS. Complete data profiles extend Table~\ref{tab:digits}.

\paragraph{Synthetic configurations on $S^2$.}
All five examples in Table~\ref{tab:synthetic} use equal weights and $n=300$ labeled atoms. The unimodal sample follows a vMF distribution on $S^2$. Its mean direction is $e_3=(0,0,1)^\top$, and its concentration is $\kappa=140$. The antipodal example combines 150 observations from this distribution with 150 from its antipodal component. The second component has mean direction $-e_3$ and concentration $\kappa=140$. For the girdle, draw $\theta\sim\mathrm{Unif}(0,2\pi)$ and $z\sim N(0,0.035^2)$. Clip $z$ to $[-0.12,0.12]$, then set
\[
  (\sqrt{1-z^2}\cos\theta,\sqrt{1-z^2}\sin\theta,z)^\top.
\]
The four-mode example uses four tetrahedral directions:
\[
  \frac{(1,1,1)}{\sqrt3},\quad
  \frac{(1,-1,-1)}{\sqrt3},\quad
  \frac{(-1,1,-1)}{\sqrt3},\quad
  \frac{(-1,-1,1)}{\sqrt3}.
\]
Each center contributes 75 vMF observations with $\kappa=140$. The duplicate example constructs 30 quasi-uniform Fibonacci-sphere locations. Repeating each location ten times gives 300 labels at 30 distinct support points. This construction tests exact atom-splitting invariance.

\paragraph{Angular scale, heat-kernel computation, and truncation.}
On $S^2$, we express diffusion time through angular resolution:
\[
  \alpha=\sqrt{8t}.
\]
We convert radians to degrees by multiplying by $180/\pi$. Under the small-time law $\exp\{-d_g(x,y)^2/(8t)\}$, distance $d_g(x,y)=\alpha$ gives leading overlap $e^{-1}$. We evaluate the heat kernel through the spectral series
\[
  k_t(x,y)=\sum_{\ell=0}^{\infty}(2\ell+1)e^{-\ell(\ell+1)t}
  P_\ell(x^\top y).
\]
The main figures and tables truncate the series at $L=180$. The replication, importance-sampling, and digit-profile figures use $L=120$, $140$, and $160$. For $k_{2t}$, the diagonal truncation tail is bounded by
\[
  \sum_{\ell>L}(2\ell+1)e^{-2t\ell(\ell+1)}.
\]
Finer diffusion scales require $L$ to increase on the order of $t^{-1/2}$. Given the heat-kernel values, evaluating $K$ scales costs $O(n^2K)$.

\paragraph{Replicated synthetic profiles.}

\begin{figure*}[t]
\centering
\input{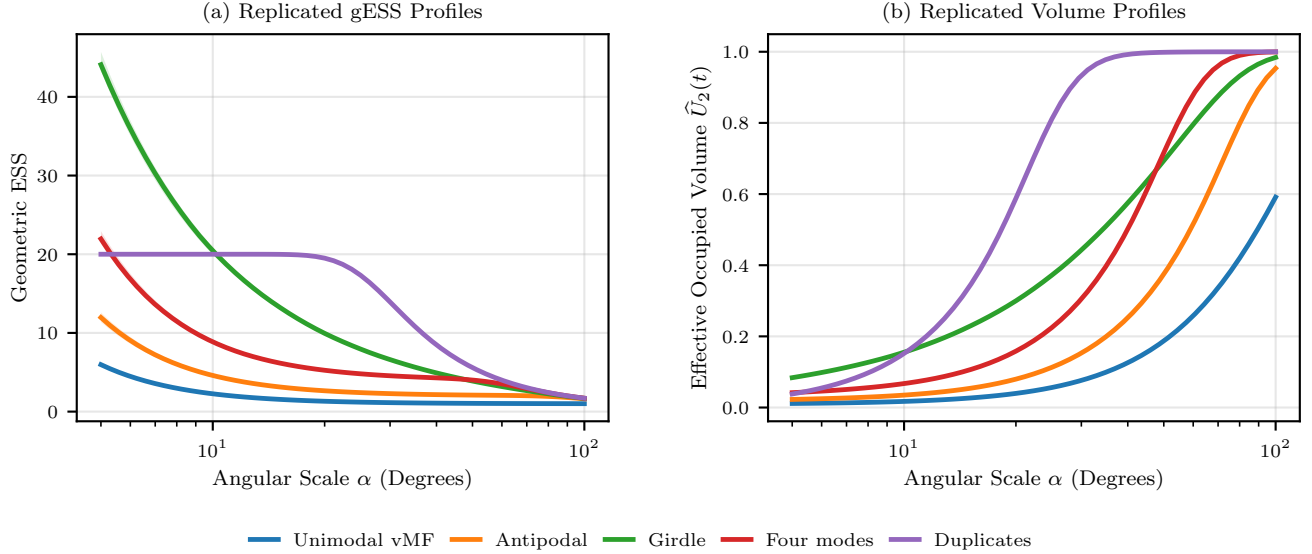}
\caption{Replicated synthetic profiles on $S^2$. Each curve is a median over 30 replications, with an interquartile band.
The left panel shows gESS. The right panel shows effective occupied volume. Profile shapes are stable across replications.}
\label{fig:app_replicates}
\end{figure*}

Figure~\ref{fig:app_replicates} uses 30 independent variants of the synthetic constructions, with $n=180$ per replication. The vMF concentration is $\kappa=100$ to make variation visible. The girdle standard deviation is $0.045$, with clipping to $[-0.15,0.15]$ before the same embedding. The duplicate example repeats each of 20 Fibonacci-sphere locations nine times.
Curves report medians and shaded interquartile ranges.

\begin{figure*}[t]
\centering
\input{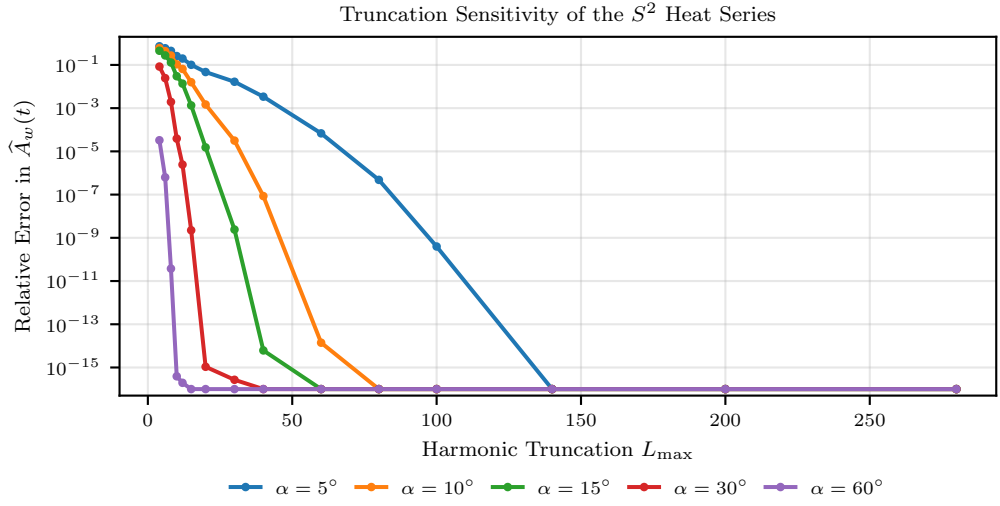}
\caption{Sensitivity of the $S^2$ heat profile to harmonic truncation. The plot compares $\widehat A_w(t)$ with a longer-series reference.
Smaller angular scales need more harmonics because heat flow has not yet damped the high frequencies.}
\label{fig:app_truncation}
\end{figure*}

\paragraph{Spectral truncation sensitivity.}
Figure~\ref{fig:app_truncation} varies $L_{\max}$ in the $S^2$ expansion. The reference uses $L_{\max}=360$. The figure reports relative error in $\widehat A_w(t)$ for a four-mode cloud. Fine angular scales require more terms, consistent with $L_{\max}$ of order $t^{-1/2}$. The main-paper cutoff is adequate for the reported scales. The finest scales are also statistically and numerically less stable.

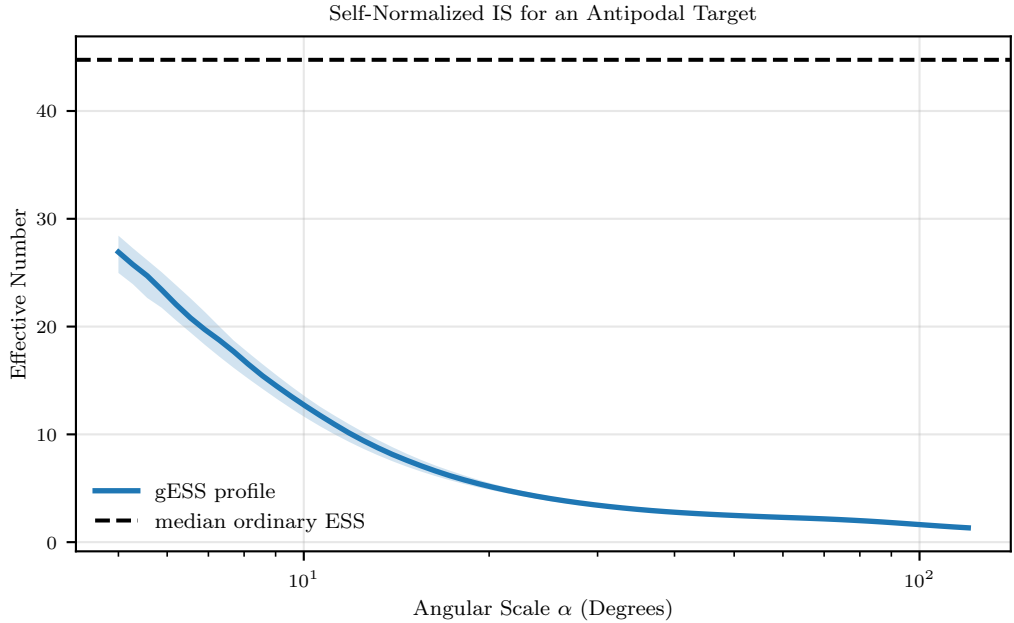
\begin{figure*}[t]
\centering
\input{figures/hkep_app_snis.pgf}
\caption{Self-normalized importance sampling for an antipodal two-mode target on $S^2$. The solid curve is median gESS over 40 trials, with an interquartile band.
The dashed line is median ordinary ESS. Only gESS shows how particle distinguishability changes with angular scale.}
\label{fig:app_snis}
\end{figure*}

\paragraph{Self-normalized importance-sampling diagnostic.}
Figure~\ref{fig:app_snis} considers self-normalized importance sampling with a uniform proposal on $S^2$. The target is an equal antipodal mixture of two vMF components. Their means are $e_3$ and $-e_3$, with $\kappa=18$. This concentration gives nontrivial importance weights and a clearly bimodal target. The density ratio is bounded because the proposal is uniform and $S^2$ is compact. Thus, Theorem~\ref{thm:consistency} applies. Ordinary ESS does not reveal whether high-weight particles occupy one region, two regions, or a diffuse set. The gESS curve adds this geometry and approaches one as the antipodal components blur together.

\paragraph{Real-data normalized embedding example.}
Table~\ref{tab:digits} uses the scikit-learn handwritten digits dataset. We standardize the $8\times8$ images featurewise and fit three principal components to the full dataset. The projected points are normalized to unit length in $\mathbb R^3$. We retain the first 30 observations of each digit. ``Digits all'' assigns equal weights to these 300 embeddings. ``Digits weighted'' uses the same points with weights proportional to $\exp(2.5x_1)$. Here, $x_1$ is the first normalized coordinate. ``Digits 0/1'' uses 60 equal-weight embeddings from digits 0 and 1. This example studies reweighting a normalized representation. It does not assume that the original images are spherical data.

\begin{figure*}[t]
\centering
\input{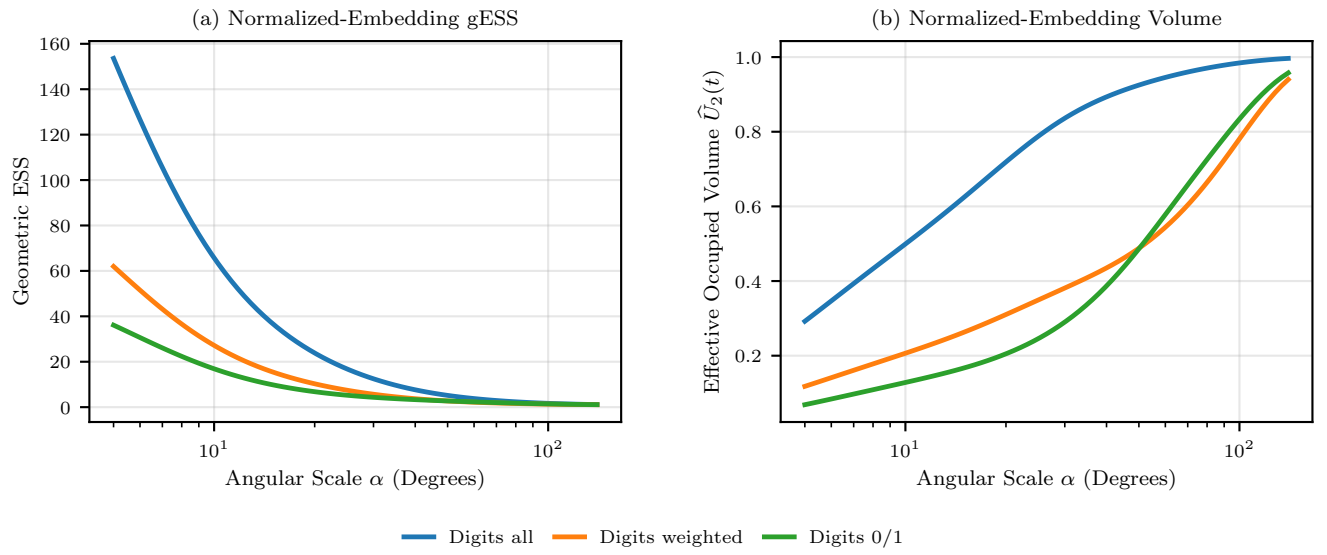}
\caption{Full profiles for the normalized embeddings in Table~\ref{tab:digits}. The left panel shows gESS, and the right shows effective occupied volume.
The weighted embedding is lower on both measures across many angular scales.}
\label{fig:app_digits_profiles}
\end{figure*}

Figure~\ref{fig:app_digits_profiles} gives the full profiles for Table~\ref{tab:digits}. The table reports $15^\circ$, while these curves cover a range of scales. The reliability-weighted embedding has lower gESS throughout much of this range. The two-class subset also differs in profile shape from the ten-class cloud. A single weight-only ESS cannot show either difference.

\end{document}

%% file: figures/hkep_app_snis.pgf
\providecommand{\mathdefault}[1]{#1}
\providecommand{\mathregular}[1]{#1}
\begingroup%
\makeatletter%
\begin{pgfpicture}%
\pgfpathrectangle{\pgfpointorigin}{\pgfqpoint{5.431158in}{3.399883in}}%
\pgfusepath{use as bounding box, clip}%
\begin{pgfscope}%
\pgfsetbuttcap%
\pgfsetmiterjoin%
\definecolor{currentfill}{rgb}{1.000000,1.000000,1.000000}%
\pgfsetfillcolor{currentfill}%
\pgfsetlinewidth{0.000000pt}%
\definecolor{currentstroke}{rgb}{1.000000,1.000000,1.000000}%
\pgfsetstrokecolor{currentstroke}%
\pgfsetdash{}{0pt}%
\pgfpathmoveto{\pgfqpoint{0.000000in}{0.000000in}}%
\pgfpathlineto{\pgfqpoint{5.431158in}{0.000000in}}%
\pgfpathlineto{\pgfqpoint{5.431158in}{3.399883in}}%
\pgfpathlineto{\pgfqpoint{0.000000in}{3.399883in}}%
\pgfpathlineto{\pgfqpoint{0.000000in}{0.000000in}}%
\pgfpathclose%
\pgfusepath{fill}%
\end{pgfscope}%
\begin{pgfscope}%
\pgfsetbuttcap%
\pgfsetmiterjoin%
\definecolor{currentfill}{rgb}{1.000000,1.000000,1.000000}%
\pgfsetfillcolor{currentfill}%
\pgfsetlinewidth{0.000000pt}%
\definecolor{currentstroke}{rgb}{0.000000,0.000000,0.000000}%
\pgfsetstrokecolor{currentstroke}%
\pgfsetstrokeopacity{0.000000}%
\pgfsetdash{}{0pt}%
\pgfpathmoveto{\pgfqpoint{0.462269in}{0.457500in}}%
\pgfpathlineto{\pgfqpoint{5.331158in}{0.457500in}}%
\pgfpathlineto{\pgfqpoint{5.331158in}{3.139389in}}%
\pgfpathlineto{\pgfqpoint{0.462269in}{3.139389in}}%
\pgfpathlineto{\pgfqpoint{0.462269in}{0.457500in}}%
\pgfpathclose%
\pgfusepath{fill}%
\end{pgfscope}%
\begin{pgfscope}%
\pgfpathrectangle{\pgfqpoint{0.462269in}{0.457500in}}{\pgfqpoint{4.868889in}{2.681889in}}%
\pgfusepath{clip}%
\pgfsetbuttcap%
\pgfsetroundjoin%
\definecolor{currentfill}{rgb}{0.121569,0.466667,0.705882}%
\pgfsetfillcolor{currentfill}%
\pgfsetfillopacity{0.200000}%
\pgfsetlinewidth{0.000000pt}%
\definecolor{currentstroke}{rgb}{0.000000,0.000000,0.000000}%
\pgfsetstrokecolor{currentstroke}%
\pgfsetdash{}{0pt}%
\pgfpathmoveto{\pgfqpoint{0.683582in}{2.101731in}}%
\pgfpathlineto{\pgfqpoint{0.683582in}{1.907344in}}%
\pgfpathlineto{\pgfqpoint{0.758604in}{1.849076in}}%
\pgfpathlineto{\pgfqpoint{0.833625in}{1.777103in}}%
\pgfpathlineto{\pgfqpoint{0.908646in}{1.726294in}}%
\pgfpathlineto{\pgfqpoint{0.983668in}{1.660214in}}%
\pgfpathlineto{\pgfqpoint{1.058689in}{1.597069in}}%
\pgfpathlineto{\pgfqpoint{1.133711in}{1.532985in}}%
\pgfpathlineto{\pgfqpoint{1.208732in}{1.471867in}}%
\pgfpathlineto{\pgfqpoint{1.283754in}{1.413162in}}%
\pgfpathlineto{\pgfqpoint{1.358775in}{1.357495in}}%
\pgfpathlineto{\pgfqpoint{1.433796in}{1.303764in}}%
\pgfpathlineto{\pgfqpoint{1.508818in}{1.251271in}}%
\pgfpathlineto{\pgfqpoint{1.583839in}{1.201105in}}%
\pgfpathlineto{\pgfqpoint{1.658861in}{1.153515in}}%
\pgfpathlineto{\pgfqpoint{1.733882in}{1.108773in}}%
\pgfpathlineto{\pgfqpoint{1.808903in}{1.066913in}}%
\pgfpathlineto{\pgfqpoint{1.883925in}{1.027922in}}%
\pgfpathlineto{\pgfqpoint{1.958946in}{0.991747in}}%
\pgfpathlineto{\pgfqpoint{2.033968in}{0.958300in}}%
\pgfpathlineto{\pgfqpoint{2.108989in}{0.927467in}}%
\pgfpathlineto{\pgfqpoint{2.184010in}{0.899117in}}%
\pgfpathlineto{\pgfqpoint{2.259032in}{0.873115in}}%
\pgfpathlineto{\pgfqpoint{2.334053in}{0.849536in}}%
\pgfpathlineto{\pgfqpoint{2.409075in}{0.827943in}}%
\pgfpathlineto{\pgfqpoint{2.484096in}{0.808199in}}%
\pgfpathlineto{\pgfqpoint{2.559117in}{0.790293in}}%
\pgfpathlineto{\pgfqpoint{2.634139in}{0.774075in}}%
\pgfpathlineto{\pgfqpoint{2.709160in}{0.759118in}}%
\pgfpathlineto{\pgfqpoint{2.784182in}{0.745013in}}%
\pgfpathlineto{\pgfqpoint{2.859203in}{0.731923in}}%
\pgfpathlineto{\pgfqpoint{2.934224in}{0.718950in}}%
\pgfpathlineto{\pgfqpoint{3.009246in}{0.708960in}}%
\pgfpathlineto{\pgfqpoint{3.084267in}{0.699903in}}%
\pgfpathlineto{\pgfqpoint{3.159289in}{0.691671in}}%
\pgfpathlineto{\pgfqpoint{3.234310in}{0.683941in}}%
\pgfpathlineto{\pgfqpoint{3.309331in}{0.676928in}}%
\pgfpathlineto{\pgfqpoint{3.384353in}{0.670592in}}%
\pgfpathlineto{\pgfqpoint{3.459374in}{0.664937in}}%
\pgfpathlineto{\pgfqpoint{3.534396in}{0.659884in}}%
\pgfpathlineto{\pgfqpoint{3.609417in}{0.655326in}}%
\pgfpathlineto{\pgfqpoint{3.684438in}{0.651212in}}%
\pgfpathlineto{\pgfqpoint{3.759460in}{0.647554in}}%
\pgfpathlineto{\pgfqpoint{3.834481in}{0.644309in}}%
\pgfpathlineto{\pgfqpoint{3.909503in}{0.640988in}}%
\pgfpathlineto{\pgfqpoint{3.984524in}{0.637813in}}%
\pgfpathlineto{\pgfqpoint{4.059545in}{0.635213in}}%
\pgfpathlineto{\pgfqpoint{4.134567in}{0.632687in}}%
\pgfpathlineto{\pgfqpoint{4.209588in}{0.630133in}}%
\pgfpathlineto{\pgfqpoint{4.284610in}{0.627550in}}%
\pgfpathlineto{\pgfqpoint{4.359631in}{0.624694in}}%
\pgfpathlineto{\pgfqpoint{4.434652in}{0.621522in}}%
\pgfpathlineto{\pgfqpoint{4.509674in}{0.617932in}}%
\pgfpathlineto{\pgfqpoint{4.584695in}{0.613867in}}%
\pgfpathlineto{\pgfqpoint{4.659717in}{0.609274in}}%
\pgfpathlineto{\pgfqpoint{4.734738in}{0.604280in}}%
\pgfpathlineto{\pgfqpoint{4.809759in}{0.599069in}}%
\pgfpathlineto{\pgfqpoint{4.884781in}{0.593811in}}%
\pgfpathlineto{\pgfqpoint{4.959802in}{0.588682in}}%
\pgfpathlineto{\pgfqpoint{5.034824in}{0.583837in}}%
\pgfpathlineto{\pgfqpoint{5.109845in}{0.579404in}}%
\pgfpathlineto{\pgfqpoint{5.109845in}{0.580153in}}%
\pgfpathlineto{\pgfqpoint{5.109845in}{0.580153in}}%
\pgfpathlineto{\pgfqpoint{5.034824in}{0.584811in}}%
\pgfpathlineto{\pgfqpoint{4.959802in}{0.589926in}}%
\pgfpathlineto{\pgfqpoint{4.884781in}{0.595334in}}%
\pgfpathlineto{\pgfqpoint{4.809759in}{0.600888in}}%
\pgfpathlineto{\pgfqpoint{4.734738in}{0.606480in}}%
\pgfpathlineto{\pgfqpoint{4.659717in}{0.611819in}}%
\pgfpathlineto{\pgfqpoint{4.584695in}{0.616747in}}%
\pgfpathlineto{\pgfqpoint{4.509674in}{0.621196in}}%
\pgfpathlineto{\pgfqpoint{4.434652in}{0.625152in}}%
\pgfpathlineto{\pgfqpoint{4.359631in}{0.628669in}}%
\pgfpathlineto{\pgfqpoint{4.284610in}{0.631850in}}%
\pgfpathlineto{\pgfqpoint{4.209588in}{0.634827in}}%
\pgfpathlineto{\pgfqpoint{4.134567in}{0.637733in}}%
\pgfpathlineto{\pgfqpoint{4.059545in}{0.640693in}}%
\pgfpathlineto{\pgfqpoint{3.984524in}{0.643801in}}%
\pgfpathlineto{\pgfqpoint{3.909503in}{0.647131in}}%
\pgfpathlineto{\pgfqpoint{3.834481in}{0.650787in}}%
\pgfpathlineto{\pgfqpoint{3.759460in}{0.654976in}}%
\pgfpathlineto{\pgfqpoint{3.684438in}{0.659486in}}%
\pgfpathlineto{\pgfqpoint{3.609417in}{0.664492in}}%
\pgfpathlineto{\pgfqpoint{3.534396in}{0.670054in}}%
\pgfpathlineto{\pgfqpoint{3.459374in}{0.676235in}}%
\pgfpathlineto{\pgfqpoint{3.384353in}{0.683102in}}%
\pgfpathlineto{\pgfqpoint{3.309331in}{0.690728in}}%
\pgfpathlineto{\pgfqpoint{3.234310in}{0.699193in}}%
\pgfpathlineto{\pgfqpoint{3.159289in}{0.708585in}}%
\pgfpathlineto{\pgfqpoint{3.084267in}{0.718997in}}%
\pgfpathlineto{\pgfqpoint{3.009246in}{0.730534in}}%
\pgfpathlineto{\pgfqpoint{2.934224in}{0.743275in}}%
\pgfpathlineto{\pgfqpoint{2.859203in}{0.757357in}}%
\pgfpathlineto{\pgfqpoint{2.784182in}{0.772903in}}%
\pgfpathlineto{\pgfqpoint{2.709160in}{0.790044in}}%
\pgfpathlineto{\pgfqpoint{2.634139in}{0.808918in}}%
\pgfpathlineto{\pgfqpoint{2.559117in}{0.829751in}}%
\pgfpathlineto{\pgfqpoint{2.484096in}{0.852854in}}%
\pgfpathlineto{\pgfqpoint{2.409075in}{0.878121in}}%
\pgfpathlineto{\pgfqpoint{2.334053in}{0.904643in}}%
\pgfpathlineto{\pgfqpoint{2.259032in}{0.934030in}}%
\pgfpathlineto{\pgfqpoint{2.184010in}{0.966130in}}%
\pgfpathlineto{\pgfqpoint{2.108989in}{1.000817in}}%
\pgfpathlineto{\pgfqpoint{2.033968in}{1.038161in}}%
\pgfpathlineto{\pgfqpoint{1.958946in}{1.078212in}}%
\pgfpathlineto{\pgfqpoint{1.883925in}{1.120362in}}%
\pgfpathlineto{\pgfqpoint{1.808903in}{1.164610in}}%
\pgfpathlineto{\pgfqpoint{1.733882in}{1.210688in}}%
\pgfpathlineto{\pgfqpoint{1.658861in}{1.261681in}}%
\pgfpathlineto{\pgfqpoint{1.583839in}{1.316443in}}%
\pgfpathlineto{\pgfqpoint{1.508818in}{1.373447in}}%
\pgfpathlineto{\pgfqpoint{1.433796in}{1.432501in}}%
\pgfpathlineto{\pgfqpoint{1.358775in}{1.493248in}}%
\pgfpathlineto{\pgfqpoint{1.283754in}{1.555735in}}%
\pgfpathlineto{\pgfqpoint{1.208732in}{1.629306in}}%
\pgfpathlineto{\pgfqpoint{1.133711in}{1.704374in}}%
\pgfpathlineto{\pgfqpoint{1.058689in}{1.775718in}}%
\pgfpathlineto{\pgfqpoint{0.983668in}{1.843875in}}%
\pgfpathlineto{\pgfqpoint{0.908646in}{1.911625in}}%
\pgfpathlineto{\pgfqpoint{0.833625in}{1.974240in}}%
\pgfpathlineto{\pgfqpoint{0.758604in}{2.036590in}}%
\pgfpathlineto{\pgfqpoint{0.683582in}{2.101731in}}%
\pgfpathlineto{\pgfqpoint{0.683582in}{2.101731in}}%
\pgfpathclose%
\pgfusepath{fill}%
\end{pgfscope}%
\begin{pgfscope}%
\pgfpathrectangle{\pgfqpoint{0.462269in}{0.457500in}}{\pgfqpoint{4.868889in}{2.681889in}}%
\pgfusepath{clip}%
\pgfsetrectcap%
\pgfsetroundjoin%
\pgfsetlinewidth{0.803000pt}%
\definecolor{currentstroke}{rgb}{0.690196,0.690196,0.690196}%
\pgfsetstrokecolor{currentstroke}%
\pgfsetstrokeopacity{0.300000}%
\pgfsetdash{}{0pt}%
\pgfpathmoveto{\pgfqpoint{1.648969in}{0.457500in}}%
\pgfpathlineto{\pgfqpoint{1.648969in}{3.139389in}}%
\pgfusepath{stroke}%
\end{pgfscope}%
\begin{pgfscope}%
\pgfsetbuttcap%
\pgfsetroundjoin%
\definecolor{currentfill}{rgb}{0.000000,0.000000,0.000000}%
\pgfsetfillcolor{currentfill}%
\pgfsetlinewidth{0.803000pt}%
\definecolor{currentstroke}{rgb}{0.000000,0.000000,0.000000}%
\pgfsetstrokecolor{currentstroke}%
\pgfsetdash{}{0pt}%
\pgfsys@defobject{currentmarker}{\pgfqpoint{0.000000in}{-0.048611in}}{\pgfqpoint{0.000000in}{0.000000in}}{%
\pgfpathmoveto{\pgfqpoint{0.000000in}{0.000000in}}%
\pgfpathlineto{\pgfqpoint{0.000000in}{-0.048611in}}%
\pgfusepath{stroke,fill}%
}%
\begin{pgfscope}%
\pgfsys@transformshift{1.648969in}{0.457500in}%
\pgfsys@useobject{currentmarker}{}%
\end{pgfscope}%
\end{pgfscope}%
\begin{pgfscope}%
\definecolor{textcolor}{rgb}{0.000000,0.000000,0.000000}%
\pgfsetstrokecolor{textcolor}%
\pgfsetfillcolor{textcolor}%
\pgftext[x=1.648969in,y=0.360278in,,top]{\color{textcolor}{\rmfamily\fontsize{7.000000}{8.400000}\selectfont\catcode`\^=\active\def^{\ifmmode\sp\else\^{}\fi}\catcode`\%=\active\def
\end{pgfscope}%
\begin{pgfscope}%
\pgfpathrectangle{\pgfqpoint{0.462269in}{0.457500in}}{\pgfqpoint{4.868889in}{2.681889in}}%
\pgfusepath{clip}%
\pgfsetrectcap%
\pgfsetroundjoin%
\pgfsetlinewidth{0.803000pt}%
\definecolor{currentstroke}{rgb}{0.690196,0.690196,0.690196}%
\pgfsetstrokecolor{currentstroke}%
\pgfsetstrokeopacity{0.300000}%
\pgfsetdash{}{0pt}%
\pgfpathmoveto{\pgfqpoint{4.855915in}{0.457500in}}%
\pgfpathlineto{\pgfqpoint{4.855915in}{3.139389in}}%
\pgfusepath{stroke}%
\end{pgfscope}%
\begin{pgfscope}%
\pgfsetbuttcap%
\pgfsetroundjoin%
\definecolor{currentfill}{rgb}{0.000000,0.000000,0.000000}%
\pgfsetfillcolor{currentfill}%
\pgfsetlinewidth{0.803000pt}%
\definecolor{currentstroke}{rgb}{0.000000,0.000000,0.000000}%
\pgfsetstrokecolor{currentstroke}%
\pgfsetdash{}{0pt}%
\pgfsys@defobject{currentmarker}{\pgfqpoint{0.000000in}{-0.048611in}}{\pgfqpoint{0.000000in}{0.000000in}}{%
\pgfpathmoveto{\pgfqpoint{0.000000in}{0.000000in}}%
\pgfpathlineto{\pgfqpoint{0.000000in}{-0.048611in}}%
\pgfusepath{stroke,fill}%
}%
\begin{pgfscope}%
\pgfsys@transformshift{4.855915in}{0.457500in}%
\pgfsys@useobject{currentmarker}{}%
\end{pgfscope}%
\end{pgfscope}%
\begin{pgfscope}%
\definecolor{textcolor}{rgb}{0.000000,0.000000,0.000000}%
\pgfsetstrokecolor{textcolor}%
\pgfsetfillcolor{textcolor}%
\pgftext[x=4.855915in,y=0.360278in,,top]{\color{textcolor}{\rmfamily\fontsize{7.000000}{8.400000}\selectfont\catcode`\^=\active\def^{\ifmmode\sp\else\^{}\fi}\catcode`\%=\active\def
\end{pgfscope}%
\begin{pgfscope}%
\pgfsetbuttcap%
\pgfsetroundjoin%
\definecolor{currentfill}{rgb}{0.000000,0.000000,0.000000}%
\pgfsetfillcolor{currentfill}%
\pgfsetlinewidth{0.602250pt}%
\definecolor{currentstroke}{rgb}{0.000000,0.000000,0.000000}%
\pgfsetstrokecolor{currentstroke}%
\pgfsetdash{}{0pt}%
\pgfsys@defobject{currentmarker}{\pgfqpoint{0.000000in}{-0.027778in}}{\pgfqpoint{0.000000in}{0.000000in}}{%
\pgfpathmoveto{\pgfqpoint{0.000000in}{0.000000in}}%
\pgfpathlineto{\pgfqpoint{0.000000in}{-0.027778in}}%
\pgfusepath{stroke,fill}%
}%
\begin{pgfscope}%
\pgfsys@transformshift{0.683582in}{0.457500in}%
\pgfsys@useobject{currentmarker}{}%
\end{pgfscope}%
\end{pgfscope}%
\begin{pgfscope}%
\pgfsetbuttcap%
\pgfsetroundjoin%
\definecolor{currentfill}{rgb}{0.000000,0.000000,0.000000}%
\pgfsetfillcolor{currentfill}%
\pgfsetlinewidth{0.602250pt}%
\definecolor{currentstroke}{rgb}{0.000000,0.000000,0.000000}%
\pgfsetstrokecolor{currentstroke}%
\pgfsetdash{}{0pt}%
\pgfsys@defobject{currentmarker}{\pgfqpoint{0.000000in}{-0.027778in}}{\pgfqpoint{0.000000in}{0.000000in}}{%
\pgfpathmoveto{\pgfqpoint{0.000000in}{0.000000in}}%
\pgfpathlineto{\pgfqpoint{0.000000in}{-0.027778in}}%
\pgfusepath{stroke,fill}%
}%
\begin{pgfscope}%
\pgfsys@transformshift{0.937512in}{0.457500in}%
\pgfsys@useobject{currentmarker}{}%
\end{pgfscope}%
\end{pgfscope}%
\begin{pgfscope}%
\pgfsetbuttcap%
\pgfsetroundjoin%
\definecolor{currentfill}{rgb}{0.000000,0.000000,0.000000}%
\pgfsetfillcolor{currentfill}%
\pgfsetlinewidth{0.602250pt}%
\definecolor{currentstroke}{rgb}{0.000000,0.000000,0.000000}%
\pgfsetstrokecolor{currentstroke}%
\pgfsetdash{}{0pt}%
\pgfsys@defobject{currentmarker}{\pgfqpoint{0.000000in}{-0.027778in}}{\pgfqpoint{0.000000in}{0.000000in}}{%
\pgfpathmoveto{\pgfqpoint{0.000000in}{0.000000in}}%
\pgfpathlineto{\pgfqpoint{0.000000in}{-0.027778in}}%
\pgfusepath{stroke,fill}%
}%
\begin{pgfscope}%
\pgfsys@transformshift{1.152207in}{0.457500in}%
\pgfsys@useobject{currentmarker}{}%
\end{pgfscope}%
\end{pgfscope}%
\begin{pgfscope}%
\pgfsetbuttcap%
\pgfsetroundjoin%
\definecolor{currentfill}{rgb}{0.000000,0.000000,0.000000}%
\pgfsetfillcolor{currentfill}%
\pgfsetlinewidth{0.602250pt}%
\definecolor{currentstroke}{rgb}{0.000000,0.000000,0.000000}%
\pgfsetstrokecolor{currentstroke}%
\pgfsetdash{}{0pt}%
\pgfsys@defobject{currentmarker}{\pgfqpoint{0.000000in}{-0.027778in}}{\pgfqpoint{0.000000in}{0.000000in}}{%
\pgfpathmoveto{\pgfqpoint{0.000000in}{0.000000in}}%
\pgfpathlineto{\pgfqpoint{0.000000in}{-0.027778in}}%
\pgfusepath{stroke,fill}%
}%
\begin{pgfscope}%
\pgfsys@transformshift{1.338184in}{0.457500in}%
\pgfsys@useobject{currentmarker}{}%
\end{pgfscope}%
\end{pgfscope}%
\begin{pgfscope}%
\pgfsetbuttcap%
\pgfsetroundjoin%
\definecolor{currentfill}{rgb}{0.000000,0.000000,0.000000}%
\pgfsetfillcolor{currentfill}%
\pgfsetlinewidth{0.602250pt}%
\definecolor{currentstroke}{rgb}{0.000000,0.000000,0.000000}%
\pgfsetstrokecolor{currentstroke}%
\pgfsetdash{}{0pt}%
\pgfsys@defobject{currentmarker}{\pgfqpoint{0.000000in}{-0.027778in}}{\pgfqpoint{0.000000in}{0.000000in}}{%
\pgfpathmoveto{\pgfqpoint{0.000000in}{0.000000in}}%
\pgfpathlineto{\pgfqpoint{0.000000in}{-0.027778in}}%
\pgfusepath{stroke,fill}%
}%
\begin{pgfscope}%
\pgfsys@transformshift{1.502227in}{0.457500in}%
\pgfsys@useobject{currentmarker}{}%
\end{pgfscope}%
\end{pgfscope}%
\begin{pgfscope}%
\pgfsetbuttcap%
\pgfsetroundjoin%
\definecolor{currentfill}{rgb}{0.000000,0.000000,0.000000}%
\pgfsetfillcolor{currentfill}%
\pgfsetlinewidth{0.602250pt}%
\definecolor{currentstroke}{rgb}{0.000000,0.000000,0.000000}%
\pgfsetstrokecolor{currentstroke}%
\pgfsetdash{}{0pt}%
\pgfsys@defobject{currentmarker}{\pgfqpoint{0.000000in}{-0.027778in}}{\pgfqpoint{0.000000in}{0.000000in}}{%
\pgfpathmoveto{\pgfqpoint{0.000000in}{0.000000in}}%
\pgfpathlineto{\pgfqpoint{0.000000in}{-0.027778in}}%
\pgfusepath{stroke,fill}%
}%
\begin{pgfscope}%
\pgfsys@transformshift{2.614356in}{0.457500in}%
\pgfsys@useobject{currentmarker}{}%
\end{pgfscope}%
\end{pgfscope}%
\begin{pgfscope}%
\pgfsetbuttcap%
\pgfsetroundjoin%
\definecolor{currentfill}{rgb}{0.000000,0.000000,0.000000}%
\pgfsetfillcolor{currentfill}%
\pgfsetlinewidth{0.602250pt}%
\definecolor{currentstroke}{rgb}{0.000000,0.000000,0.000000}%
\pgfsetstrokecolor{currentstroke}%
\pgfsetdash{}{0pt}%
\pgfsys@defobject{currentmarker}{\pgfqpoint{0.000000in}{-0.027778in}}{\pgfqpoint{0.000000in}{0.000000in}}{%
\pgfpathmoveto{\pgfqpoint{0.000000in}{0.000000in}}%
\pgfpathlineto{\pgfqpoint{0.000000in}{-0.027778in}}%
\pgfusepath{stroke,fill}%
}%
\begin{pgfscope}%
\pgfsys@transformshift{3.179071in}{0.457500in}%
\pgfsys@useobject{currentmarker}{}%
\end{pgfscope}%
\end{pgfscope}%
\begin{pgfscope}%
\pgfsetbuttcap%
\pgfsetroundjoin%
\definecolor{currentfill}{rgb}{0.000000,0.000000,0.000000}%
\pgfsetfillcolor{currentfill}%
\pgfsetlinewidth{0.602250pt}%
\definecolor{currentstroke}{rgb}{0.000000,0.000000,0.000000}%
\pgfsetstrokecolor{currentstroke}%
\pgfsetdash{}{0pt}%
\pgfsys@defobject{currentmarker}{\pgfqpoint{0.000000in}{-0.027778in}}{\pgfqpoint{0.000000in}{0.000000in}}{%
\pgfpathmoveto{\pgfqpoint{0.000000in}{0.000000in}}%
\pgfpathlineto{\pgfqpoint{0.000000in}{-0.027778in}}%
\pgfusepath{stroke,fill}%
}%
\begin{pgfscope}%
\pgfsys@transformshift{3.579743in}{0.457500in}%
\pgfsys@useobject{currentmarker}{}%
\end{pgfscope}%
\end{pgfscope}%
\begin{pgfscope}%
\pgfsetbuttcap%
\pgfsetroundjoin%
\definecolor{currentfill}{rgb}{0.000000,0.000000,0.000000}%
\pgfsetfillcolor{currentfill}%
\pgfsetlinewidth{0.602250pt}%
\definecolor{currentstroke}{rgb}{0.000000,0.000000,0.000000}%
\pgfsetstrokecolor{currentstroke}%
\pgfsetdash{}{0pt}%
\pgfsys@defobject{currentmarker}{\pgfqpoint{0.000000in}{-0.027778in}}{\pgfqpoint{0.000000in}{0.000000in}}{%
\pgfpathmoveto{\pgfqpoint{0.000000in}{0.000000in}}%
\pgfpathlineto{\pgfqpoint{0.000000in}{-0.027778in}}%
\pgfusepath{stroke,fill}%
}%
\begin{pgfscope}%
\pgfsys@transformshift{3.890528in}{0.457500in}%
\pgfsys@useobject{currentmarker}{}%
\end{pgfscope}%
\end{pgfscope}%
\begin{pgfscope}%
\pgfsetbuttcap%
\pgfsetroundjoin%
\definecolor{currentfill}{rgb}{0.000000,0.000000,0.000000}%
\pgfsetfillcolor{currentfill}%
\pgfsetlinewidth{0.602250pt}%
\definecolor{currentstroke}{rgb}{0.000000,0.000000,0.000000}%
\pgfsetstrokecolor{currentstroke}%
\pgfsetdash{}{0pt}%
\pgfsys@defobject{currentmarker}{\pgfqpoint{0.000000in}{-0.027778in}}{\pgfqpoint{0.000000in}{0.000000in}}{%
\pgfpathmoveto{\pgfqpoint{0.000000in}{0.000000in}}%
\pgfpathlineto{\pgfqpoint{0.000000in}{-0.027778in}}%
\pgfusepath{stroke,fill}%
}%
\begin{pgfscope}%
\pgfsys@transformshift{4.144458in}{0.457500in}%
\pgfsys@useobject{currentmarker}{}%
\end{pgfscope}%
\end{pgfscope}%
\begin{pgfscope}%
\pgfsetbuttcap%
\pgfsetroundjoin%
\definecolor{currentfill}{rgb}{0.000000,0.000000,0.000000}%
\pgfsetfillcolor{currentfill}%
\pgfsetlinewidth{0.602250pt}%
\definecolor{currentstroke}{rgb}{0.000000,0.000000,0.000000}%
\pgfsetstrokecolor{currentstroke}%
\pgfsetdash{}{0pt}%
\pgfsys@defobject{currentmarker}{\pgfqpoint{0.000000in}{-0.027778in}}{\pgfqpoint{0.000000in}{0.000000in}}{%
\pgfpathmoveto{\pgfqpoint{0.000000in}{0.000000in}}%
\pgfpathlineto{\pgfqpoint{0.000000in}{-0.027778in}}%
\pgfusepath{stroke,fill}%
}%
\begin{pgfscope}%
\pgfsys@transformshift{4.359153in}{0.457500in}%
\pgfsys@useobject{currentmarker}{}%
\end{pgfscope}%
\end{pgfscope}%
\begin{pgfscope}%
\pgfsetbuttcap%
\pgfsetroundjoin%
\definecolor{currentfill}{rgb}{0.000000,0.000000,0.000000}%
\pgfsetfillcolor{currentfill}%
\pgfsetlinewidth{0.602250pt}%
\definecolor{currentstroke}{rgb}{0.000000,0.000000,0.000000}%
\pgfsetstrokecolor{currentstroke}%
\pgfsetdash{}{0pt}%
\pgfsys@defobject{currentmarker}{\pgfqpoint{0.000000in}{-0.027778in}}{\pgfqpoint{0.000000in}{0.000000in}}{%
\pgfpathmoveto{\pgfqpoint{0.000000in}{0.000000in}}%
\pgfpathlineto{\pgfqpoint{0.000000in}{-0.027778in}}%
\pgfusepath{stroke,fill}%
}%
\begin{pgfscope}%
\pgfsys@transformshift{4.545130in}{0.457500in}%
\pgfsys@useobject{currentmarker}{}%
\end{pgfscope}%
\end{pgfscope}%
\begin{pgfscope}%
\pgfsetbuttcap%
\pgfsetroundjoin%
\definecolor{currentfill}{rgb}{0.000000,0.000000,0.000000}%
\pgfsetfillcolor{currentfill}%
\pgfsetlinewidth{0.602250pt}%
\definecolor{currentstroke}{rgb}{0.000000,0.000000,0.000000}%
\pgfsetstrokecolor{currentstroke}%
\pgfsetdash{}{0pt}%
\pgfsys@defobject{currentmarker}{\pgfqpoint{0.000000in}{-0.027778in}}{\pgfqpoint{0.000000in}{0.000000in}}{%
\pgfpathmoveto{\pgfqpoint{0.000000in}{0.000000in}}%
\pgfpathlineto{\pgfqpoint{0.000000in}{-0.027778in}}%
\pgfusepath{stroke,fill}%
}%
\begin{pgfscope}%
\pgfsys@transformshift{4.709173in}{0.457500in}%
\pgfsys@useobject{currentmarker}{}%
\end{pgfscope}%
\end{pgfscope}%
\begin{pgfscope}%
\definecolor{textcolor}{rgb}{0.000000,0.000000,0.000000}%
\pgfsetstrokecolor{textcolor}%
\pgfsetfillcolor{textcolor}%
\pgftext[x=2.896714in,y=0.211111in,,top]{\color{textcolor}{\rmfamily\fontsize{8.000000}{9.600000}\selectfont\catcode`\^=\active\def^{\ifmmode\sp\else\^{}\fi}\catcode`\%=\active\def
\end{pgfscope}%
\begin{pgfscope}%
\pgfpathrectangle{\pgfqpoint{0.462269in}{0.457500in}}{\pgfqpoint{4.868889in}{2.681889in}}%
\pgfusepath{clip}%
\pgfsetrectcap%
\pgfsetroundjoin%
\pgfsetlinewidth{0.803000pt}%
\definecolor{currentstroke}{rgb}{0.690196,0.690196,0.690196}%
\pgfsetstrokecolor{currentstroke}%
\pgfsetstrokeopacity{0.300000}%
\pgfsetdash{}{0pt}%
\pgfpathmoveto{\pgfqpoint{0.462269in}{0.505289in}}%
\pgfpathlineto{\pgfqpoint{5.331158in}{0.505289in}}%
\pgfusepath{stroke}%
\end{pgfscope}%
\begin{pgfscope}%
\pgfsetbuttcap%
\pgfsetroundjoin%
\definecolor{currentfill}{rgb}{0.000000,0.000000,0.000000}%
\pgfsetfillcolor{currentfill}%
\pgfsetlinewidth{0.803000pt}%
\definecolor{currentstroke}{rgb}{0.000000,0.000000,0.000000}%
\pgfsetstrokecolor{currentstroke}%
\pgfsetdash{}{0pt}%
\pgfsys@defobject{currentmarker}{\pgfqpoint{-0.048611in}{0.000000in}}{\pgfqpoint{-0.000000in}{0.000000in}}{%
\pgfpathmoveto{\pgfqpoint{-0.000000in}{0.000000in}}%
\pgfpathlineto{\pgfqpoint{-0.048611in}{0.000000in}}%
\pgfusepath{stroke,fill}%
}%
\begin{pgfscope}%
\pgfsys@transformshift{0.462269in}{0.505289in}%
\pgfsys@useobject{currentmarker}{}%
\end{pgfscope}%
\end{pgfscope}%
\begin{pgfscope}%
\definecolor{textcolor}{rgb}{0.000000,0.000000,0.000000}%
\pgfsetstrokecolor{textcolor}%
\pgfsetfillcolor{textcolor}%
\pgftext[x=0.309684in, y=0.471532in, left, base]{\color{textcolor}{\rmfamily\fontsize{7.000000}{8.400000}\selectfont\catcode`\^=\active\def^{\ifmmode\sp\else\^{}\fi}\catcode`\%=\active\def
\end{pgfscope}%
\begin{pgfscope}%
\pgfpathrectangle{\pgfqpoint{0.462269in}{0.457500in}}{\pgfqpoint{4.868889in}{2.681889in}}%
\pgfusepath{clip}%
\pgfsetrectcap%
\pgfsetroundjoin%
\pgfsetlinewidth{0.803000pt}%
\definecolor{currentstroke}{rgb}{0.690196,0.690196,0.690196}%
\pgfsetstrokecolor{currentstroke}%
\pgfsetstrokeopacity{0.300000}%
\pgfsetdash{}{0pt}%
\pgfpathmoveto{\pgfqpoint{0.462269in}{1.066683in}}%
\pgfpathlineto{\pgfqpoint{5.331158in}{1.066683in}}%
\pgfusepath{stroke}%
\end{pgfscope}%
\begin{pgfscope}%
\pgfsetbuttcap%
\pgfsetroundjoin%
\definecolor{currentfill}{rgb}{0.000000,0.000000,0.000000}%
\pgfsetfillcolor{currentfill}%
\pgfsetlinewidth{0.803000pt}%
\definecolor{currentstroke}{rgb}{0.000000,0.000000,0.000000}%
\pgfsetstrokecolor{currentstroke}%
\pgfsetdash{}{0pt}%
\pgfsys@defobject{currentmarker}{\pgfqpoint{-0.048611in}{0.000000in}}{\pgfqpoint{-0.000000in}{0.000000in}}{%
\pgfpathmoveto{\pgfqpoint{-0.000000in}{0.000000in}}%
\pgfpathlineto{\pgfqpoint{-0.048611in}{0.000000in}}%
\pgfusepath{stroke,fill}%
}%
\begin{pgfscope}%
\pgfsys@transformshift{0.462269in}{1.066683in}%
\pgfsys@useobject{currentmarker}{}%
\end{pgfscope}%
\end{pgfscope}%
\begin{pgfscope}%
\definecolor{textcolor}{rgb}{0.000000,0.000000,0.000000}%
\pgfsetstrokecolor{textcolor}%
\pgfsetfillcolor{textcolor}%
\pgftext[x=0.254321in, y=1.032926in, left, base]{\color{textcolor}{\rmfamily\fontsize{7.000000}{8.400000}\selectfont\catcode`\^=\active\def^{\ifmmode\sp\else\^{}\fi}\catcode`\%=\active\def
\end{pgfscope}%
\begin{pgfscope}%
\pgfpathrectangle{\pgfqpoint{0.462269in}{0.457500in}}{\pgfqpoint{4.868889in}{2.681889in}}%
\pgfusepath{clip}%
\pgfsetrectcap%
\pgfsetroundjoin%
\pgfsetlinewidth{0.803000pt}%
\definecolor{currentstroke}{rgb}{0.690196,0.690196,0.690196}%
\pgfsetstrokecolor{currentstroke}%
\pgfsetstrokeopacity{0.300000}%
\pgfsetdash{}{0pt}%
\pgfpathmoveto{\pgfqpoint{0.462269in}{1.628077in}}%
\pgfpathlineto{\pgfqpoint{5.331158in}{1.628077in}}%
\pgfusepath{stroke}%
\end{pgfscope}%
\begin{pgfscope}%
\pgfsetbuttcap%
\pgfsetroundjoin%
\definecolor{currentfill}{rgb}{0.000000,0.000000,0.000000}%
\pgfsetfillcolor{currentfill}%
\pgfsetlinewidth{0.803000pt}%
\definecolor{currentstroke}{rgb}{0.000000,0.000000,0.000000}%
\pgfsetstrokecolor{currentstroke}%
\pgfsetdash{}{0pt}%
\pgfsys@defobject{currentmarker}{\pgfqpoint{-0.048611in}{0.000000in}}{\pgfqpoint{-0.000000in}{0.000000in}}{%
\pgfpathmoveto{\pgfqpoint{-0.000000in}{0.000000in}}%
\pgfpathlineto{\pgfqpoint{-0.048611in}{0.000000in}}%
\pgfusepath{stroke,fill}%
}%
\begin{pgfscope}%
\pgfsys@transformshift{0.462269in}{1.628077in}%
\pgfsys@useobject{currentmarker}{}%
\end{pgfscope}%
\end{pgfscope}%
\begin{pgfscope}%
\definecolor{textcolor}{rgb}{0.000000,0.000000,0.000000}%
\pgfsetstrokecolor{textcolor}%
\pgfsetfillcolor{textcolor}%
\pgftext[x=0.254321in, y=1.594320in, left, base]{\color{textcolor}{\rmfamily\fontsize{7.000000}{8.400000}\selectfont\catcode`\^=\active\def^{\ifmmode\sp\else\^{}\fi}\catcode`\%=\active\def
\end{pgfscope}%
\begin{pgfscope}%
\pgfpathrectangle{\pgfqpoint{0.462269in}{0.457500in}}{\pgfqpoint{4.868889in}{2.681889in}}%
\pgfusepath{clip}%
\pgfsetrectcap%
\pgfsetroundjoin%
\pgfsetlinewidth{0.803000pt}%
\definecolor{currentstroke}{rgb}{0.690196,0.690196,0.690196}%
\pgfsetstrokecolor{currentstroke}%
\pgfsetstrokeopacity{0.300000}%
\pgfsetdash{}{0pt}%
\pgfpathmoveto{\pgfqpoint{0.462269in}{2.189472in}}%
\pgfpathlineto{\pgfqpoint{5.331158in}{2.189472in}}%
\pgfusepath{stroke}%
\end{pgfscope}%
\begin{pgfscope}%
\pgfsetbuttcap%
\pgfsetroundjoin%
\definecolor{currentfill}{rgb}{0.000000,0.000000,0.000000}%
\pgfsetfillcolor{currentfill}%
\pgfsetlinewidth{0.803000pt}%
\definecolor{currentstroke}{rgb}{0.000000,0.000000,0.000000}%
\pgfsetstrokecolor{currentstroke}%
\pgfsetdash{}{0pt}%
\pgfsys@defobject{currentmarker}{\pgfqpoint{-0.048611in}{0.000000in}}{\pgfqpoint{-0.000000in}{0.000000in}}{%
\pgfpathmoveto{\pgfqpoint{-0.000000in}{0.000000in}}%
\pgfpathlineto{\pgfqpoint{-0.048611in}{0.000000in}}%
\pgfusepath{stroke,fill}%
}%
\begin{pgfscope}%
\pgfsys@transformshift{0.462269in}{2.189472in}%
\pgfsys@useobject{currentmarker}{}%
\end{pgfscope}%
\end{pgfscope}%
\begin{pgfscope}%
\definecolor{textcolor}{rgb}{0.000000,0.000000,0.000000}%
\pgfsetstrokecolor{textcolor}%
\pgfsetfillcolor{textcolor}%
\pgftext[x=0.254321in, y=2.155714in, left, base]{\color{textcolor}{\rmfamily\fontsize{7.000000}{8.400000}\selectfont\catcode`\^=\active\def^{\ifmmode\sp\else\^{}\fi}\catcode`\%=\active\def
\end{pgfscope}%
\begin{pgfscope}%
\pgfpathrectangle{\pgfqpoint{0.462269in}{0.457500in}}{\pgfqpoint{4.868889in}{2.681889in}}%
\pgfusepath{clip}%
\pgfsetrectcap%
\pgfsetroundjoin%
\pgfsetlinewidth{0.803000pt}%
\definecolor{currentstroke}{rgb}{0.690196,0.690196,0.690196}%
\pgfsetstrokecolor{currentstroke}%
\pgfsetstrokeopacity{0.300000}%
\pgfsetdash{}{0pt}%
\pgfpathmoveto{\pgfqpoint{0.462269in}{2.750866in}}%
\pgfpathlineto{\pgfqpoint{5.331158in}{2.750866in}}%
\pgfusepath{stroke}%
\end{pgfscope}%
\begin{pgfscope}%
\pgfsetbuttcap%
\pgfsetroundjoin%
\definecolor{currentfill}{rgb}{0.000000,0.000000,0.000000}%
\pgfsetfillcolor{currentfill}%
\pgfsetlinewidth{0.803000pt}%
\definecolor{currentstroke}{rgb}{0.000000,0.000000,0.000000}%
\pgfsetstrokecolor{currentstroke}%
\pgfsetdash{}{0pt}%
\pgfsys@defobject{currentmarker}{\pgfqpoint{-0.048611in}{0.000000in}}{\pgfqpoint{-0.000000in}{0.000000in}}{%
\pgfpathmoveto{\pgfqpoint{-0.000000in}{0.000000in}}%
\pgfpathlineto{\pgfqpoint{-0.048611in}{0.000000in}}%
\pgfusepath{stroke,fill}%
}%
\begin{pgfscope}%
\pgfsys@transformshift{0.462269in}{2.750866in}%
\pgfsys@useobject{currentmarker}{}%
\end{pgfscope}%
\end{pgfscope}%
\begin{pgfscope}%
\definecolor{textcolor}{rgb}{0.000000,0.000000,0.000000}%
\pgfsetstrokecolor{textcolor}%
\pgfsetfillcolor{textcolor}%
\pgftext[x=0.254321in, y=2.717108in, left, base]{\color{textcolor}{\rmfamily\fontsize{7.000000}{8.400000}\selectfont\catcode`\^=\active\def^{\ifmmode\sp\else\^{}\fi}\catcode`\%=\active\def
\end{pgfscope}%
\begin{pgfscope}%
\definecolor{textcolor}{rgb}{0.000000,0.000000,0.000000}%
\pgfsetstrokecolor{textcolor}%
\pgfsetfillcolor{textcolor}%
\pgftext[x=0.198766in,y=1.798444in,,bottom,rotate=90.000000]{\color{textcolor}{\rmfamily\fontsize{8.000000}{9.600000}\selectfont\catcode`\^=\active\def^{\ifmmode\sp\else\^{}\fi}\catcode`\%=\active\def
\end{pgfscope}%
\begin{pgfscope}%
\pgfpathrectangle{\pgfqpoint{0.462269in}{0.457500in}}{\pgfqpoint{4.868889in}{2.681889in}}%
\pgfusepath{clip}%
\pgfsetrectcap%
\pgfsetroundjoin%
\pgfsetlinewidth{2.007500pt}%
\definecolor{currentstroke}{rgb}{0.121569,0.466667,0.705882}%
\pgfsetstrokecolor{currentstroke}%
\pgfsetdash{}{0pt}%
\pgfpathmoveto{\pgfqpoint{0.683582in}{2.015789in}}%
\pgfpathlineto{\pgfqpoint{0.758604in}{1.951216in}}%
\pgfpathlineto{\pgfqpoint{0.833625in}{1.892504in}}%
\pgfpathlineto{\pgfqpoint{0.908646in}{1.819530in}}%
\pgfpathlineto{\pgfqpoint{0.983668in}{1.743327in}}%
\pgfpathlineto{\pgfqpoint{1.058689in}{1.673211in}}%
\pgfpathlineto{\pgfqpoint{1.133711in}{1.612457in}}%
\pgfpathlineto{\pgfqpoint{1.208732in}{1.557837in}}%
\pgfpathlineto{\pgfqpoint{1.283754in}{1.498185in}}%
\pgfpathlineto{\pgfqpoint{1.358775in}{1.433108in}}%
\pgfpathlineto{\pgfqpoint{1.433796in}{1.372000in}}%
\pgfpathlineto{\pgfqpoint{1.508818in}{1.317106in}}%
\pgfpathlineto{\pgfqpoint{1.583839in}{1.264229in}}%
\pgfpathlineto{\pgfqpoint{1.658861in}{1.213545in}}%
\pgfpathlineto{\pgfqpoint{1.733882in}{1.165205in}}%
\pgfpathlineto{\pgfqpoint{1.808903in}{1.118892in}}%
\pgfpathlineto{\pgfqpoint{1.883925in}{1.074387in}}%
\pgfpathlineto{\pgfqpoint{1.958946in}{1.034217in}}%
\pgfpathlineto{\pgfqpoint{2.033968in}{0.997056in}}%
\pgfpathlineto{\pgfqpoint{2.108989in}{0.962755in}}%
\pgfpathlineto{\pgfqpoint{2.184010in}{0.932051in}}%
\pgfpathlineto{\pgfqpoint{2.259032in}{0.903067in}}%
\pgfpathlineto{\pgfqpoint{2.334053in}{0.876267in}}%
\pgfpathlineto{\pgfqpoint{2.409075in}{0.851613in}}%
\pgfpathlineto{\pgfqpoint{2.484096in}{0.829634in}}%
\pgfpathlineto{\pgfqpoint{2.559117in}{0.809576in}}%
\pgfpathlineto{\pgfqpoint{2.634139in}{0.791197in}}%
\pgfpathlineto{\pgfqpoint{2.709160in}{0.774378in}}%
\pgfpathlineto{\pgfqpoint{2.784182in}{0.759047in}}%
\pgfpathlineto{\pgfqpoint{2.859203in}{0.744906in}}%
\pgfpathlineto{\pgfqpoint{2.934224in}{0.732031in}}%
\pgfpathlineto{\pgfqpoint{3.009246in}{0.720335in}}%
\pgfpathlineto{\pgfqpoint{3.084267in}{0.709722in}}%
\pgfpathlineto{\pgfqpoint{3.159289in}{0.700101in}}%
\pgfpathlineto{\pgfqpoint{3.234310in}{0.691386in}}%
\pgfpathlineto{\pgfqpoint{3.309331in}{0.683498in}}%
\pgfpathlineto{\pgfqpoint{3.384353in}{0.676318in}}%
\pgfpathlineto{\pgfqpoint{3.459374in}{0.669878in}}%
\pgfpathlineto{\pgfqpoint{3.534396in}{0.664243in}}%
\pgfpathlineto{\pgfqpoint{3.609417in}{0.659240in}}%
\pgfpathlineto{\pgfqpoint{3.684438in}{0.654722in}}%
\pgfpathlineto{\pgfqpoint{3.759460in}{0.650639in}}%
\pgfpathlineto{\pgfqpoint{3.834481in}{0.646944in}}%
\pgfpathlineto{\pgfqpoint{3.909503in}{0.643587in}}%
\pgfpathlineto{\pgfqpoint{3.984524in}{0.640414in}}%
\pgfpathlineto{\pgfqpoint{4.059545in}{0.637418in}}%
\pgfpathlineto{\pgfqpoint{4.134567in}{0.634671in}}%
\pgfpathlineto{\pgfqpoint{4.209588in}{0.632100in}}%
\pgfpathlineto{\pgfqpoint{4.284610in}{0.629411in}}%
\pgfpathlineto{\pgfqpoint{4.359631in}{0.626494in}}%
\pgfpathlineto{\pgfqpoint{4.434652in}{0.623128in}}%
\pgfpathlineto{\pgfqpoint{4.509674in}{0.619282in}}%
\pgfpathlineto{\pgfqpoint{4.584695in}{0.615158in}}%
\pgfpathlineto{\pgfqpoint{4.659717in}{0.610538in}}%
\pgfpathlineto{\pgfqpoint{4.734738in}{0.605443in}}%
\pgfpathlineto{\pgfqpoint{4.809759in}{0.600108in}}%
\pgfpathlineto{\pgfqpoint{4.884781in}{0.594725in}}%
\pgfpathlineto{\pgfqpoint{4.959802in}{0.589450in}}%
\pgfpathlineto{\pgfqpoint{5.034824in}{0.584458in}}%
\pgfpathlineto{\pgfqpoint{5.109845in}{0.579888in}}%
\pgfusepath{stroke}%
\end{pgfscope}%
\begin{pgfscope}%
\pgfpathrectangle{\pgfqpoint{0.462269in}{0.457500in}}{\pgfqpoint{4.868889in}{2.681889in}}%
\pgfusepath{clip}%
\pgfsetbuttcap%
\pgfsetroundjoin%
\pgfsetlinewidth{1.505625pt}%
\definecolor{currentstroke}{rgb}{0.000000,0.000000,0.000000}%
\pgfsetstrokecolor{currentstroke}%
\pgfsetdash{{5.550000pt}{2.400000pt}}{0.000000pt}%
\pgfpathmoveto{\pgfqpoint{0.462269in}{3.017485in}}%
\pgfpathlineto{\pgfqpoint{5.331158in}{3.017485in}}%
\pgfusepath{stroke}%
\end{pgfscope}%
\begin{pgfscope}%
\pgfsetrectcap%
\pgfsetmiterjoin%
\pgfsetlinewidth{0.803000pt}%
\definecolor{currentstroke}{rgb}{0.000000,0.000000,0.000000}%
\pgfsetstrokecolor{currentstroke}%
\pgfsetdash{}{0pt}%
\pgfpathmoveto{\pgfqpoint{0.462269in}{0.457500in}}%
\pgfpathlineto{\pgfqpoint{0.462269in}{3.139389in}}%
\pgfusepath{stroke}%
\end{pgfscope}%
\begin{pgfscope}%
\pgfsetrectcap%
\pgfsetmiterjoin%
\pgfsetlinewidth{0.803000pt}%
\definecolor{currentstroke}{rgb}{0.000000,0.000000,0.000000}%
\pgfsetstrokecolor{currentstroke}%
\pgfsetdash{}{0pt}%
\pgfpathmoveto{\pgfqpoint{5.331158in}{0.457500in}}%
\pgfpathlineto{\pgfqpoint{5.331158in}{3.139389in}}%
\pgfusepath{stroke}%
\end{pgfscope}%
\begin{pgfscope}%
\pgfsetrectcap%
\pgfsetmiterjoin%
\pgfsetlinewidth{0.803000pt}%
\definecolor{currentstroke}{rgb}{0.000000,0.000000,0.000000}%
\pgfsetstrokecolor{currentstroke}%
\pgfsetdash{}{0pt}%
\pgfpathmoveto{\pgfqpoint{0.462269in}{0.457500in}}%
\pgfpathlineto{\pgfqpoint{5.331158in}{0.457500in}}%
\pgfusepath{stroke}%
\end{pgfscope}%
\begin{pgfscope}%
\pgfsetrectcap%
\pgfsetmiterjoin%
\pgfsetlinewidth{0.803000pt}%
\definecolor{currentstroke}{rgb}{0.000000,0.000000,0.000000}%
\pgfsetstrokecolor{currentstroke}%
\pgfsetdash{}{0pt}%
\pgfpathmoveto{\pgfqpoint{0.462269in}{3.139389in}}%
\pgfpathlineto{\pgfqpoint{5.331158in}{3.139389in}}%
\pgfusepath{stroke}%
\end{pgfscope}%
\begin{pgfscope}%
\definecolor{textcolor}{rgb}{0.000000,0.000000,0.000000}%
\pgfsetstrokecolor{textcolor}%
\pgfsetfillcolor{textcolor}%
\pgftext[x=2.896714in,y=3.222722in,,base]{\color{textcolor}{\rmfamily\fontsize{8.000000}{9.600000}\selectfont\catcode`\^=\active\def^{\ifmmode\sp\else\^{}\fi}\catcode`\%=\active\def
\end{pgfscope}%
\begin{pgfscope}%
\pgfsetrectcap%
\pgfsetroundjoin%
\pgfsetlinewidth{2.007500pt}%
\definecolor{currentstroke}{rgb}{0.121569,0.466667,0.705882}%
\pgfsetstrokecolor{currentstroke}%
\pgfsetdash{}{0pt}%
\pgfpathmoveto{\pgfqpoint{0.562269in}{0.772932in}}%
\pgfpathlineto{\pgfqpoint{0.673380in}{0.772932in}}%
\pgfpathlineto{\pgfqpoint{0.784491in}{0.772932in}}%
\pgfusepath{stroke}%
\end{pgfscope}%
\begin{pgfscope}%
\definecolor{textcolor}{rgb}{0.000000,0.000000,0.000000}%
\pgfsetstrokecolor{textcolor}%
\pgfsetfillcolor{textcolor}%
\pgftext[x=0.873380in,y=0.734043in,left,base]{\color{textcolor}{\rmfamily\fontsize{8.000000}{9.600000}\selectfont\catcode`\^=\active\def^{\ifmmode\sp\else\^{}\fi}\catcode`\%=\active\def
\end{pgfscope}%
\begin{pgfscope}%
\pgfsetbuttcap%
\pgfsetroundjoin%
\pgfsetlinewidth{1.505625pt}%
\definecolor{currentstroke}{rgb}{0.000000,0.000000,0.000000}%
\pgfsetstrokecolor{currentstroke}%
\pgfsetdash{{5.550000pt}{2.400000pt}}{0.000000pt}%
\pgfpathmoveto{\pgfqpoint{0.562269in}{0.617994in}}%
\pgfpathlineto{\pgfqpoint{0.673380in}{0.617994in}}%
\pgfpathlineto{\pgfqpoint{0.784491in}{0.617994in}}%
\pgfusepath{stroke}%
\end{pgfscope}%
\begin{pgfscope}%
\definecolor{textcolor}{rgb}{0.000000,0.000000,0.000000}%
\pgfsetstrokecolor{textcolor}%
\pgfsetfillcolor{textcolor}%
\pgftext[x=0.873380in,y=0.579105in,left,base]{\color{textcolor}{\rmfamily\fontsize{8.000000}{9.600000}\selectfont\catcode`\^=\active\def^{\ifmmode\sp\else\^{}\fi}\catcode`\%=\active\def
\end{pgfscope}%
\end{pgfpicture}%
\makeatother%
\endgroup%

%% file: references.bib
@article{gine_1975_InvariantTestsUniformity,
	title = {Invariant {Tests} for {Uniformity} on {Compact} {Riemannian} {Manifolds} {Based} on {Sobolev} {Norms}},
	volume = {3},
	issn = {0090-5364},
	doi = {10.1214/aos/1176343283},
	number = {6},
	urldate = {2026-07-04},
	journal = {The Annals of Statistics},
	author = {Gine, Evarist},
	month = nov,
	year = {1975},
}

@article{sriperumbudur_2010_HilbertSpaceEmbeddings,
	title = {Hilbert {Space} {Embeddings} and {Metrics} on {Probability} {Measures}},
	volume = {11},
	issn = {1532-4435},
	journal = {Journal of Machine Learning Research},
	publisher = {JMLR.org},
	author = {Sriperumbudur, Bharath K. and Gretton, Arthur and Fukumizu, Kenji and Schölkopf, Bernhard and Lanckriet, Gert R.G.},
	month = aug,
	year = {2010},
	pages = {1517--1561},
}

@article{liu_1998_SequentialMonteCarlo,
	title = {Sequential {Monte} {Carlo} {Methods} for {Dynamic} {Systems}},
	volume = {93},
	issn = {0162-1459, 1537-274X},
	doi = {10.1080/01621459.1998.10473765},
	language = {en},
	number = {443},
	urldate = {2026-07-04},
	journal = {Journal of the American Statistical Association},
	author = {Liu, Jun S. and Chen, Rong},
	month = sep,
	year = {1998},
	pages = {1032--1044},
}

@article{kong_1994_SequentialImputationsBayesian,
	title = {Sequential {Imputations} and {Bayesian} {Missing} {Data} {Problems}},
	volume = {89},
	issn = {0162-1459, 1537-274X},
	doi = {10.1080/01621459.1994.10476469},
	language = {en},
	number = {425},
	urldate = {2026-07-04},
	journal = {Journal of the American Statistical Association},
	author = {Kong, Augustine and Liu, Jun S. and Wong, Wing Hung},
	month = mar,
	year = {1994},
	pages = {278--288},
}

@article{delmoral_2006_SequentialMonteCarlo,
	title = {Sequential {Monte} {Carlo} {Samplers}},
	volume = {68},
	issn = {1369-7412, 1467-9868},
	doi = {10.1111/j.1467-9868.2006.00553.x},
	language = {en},
	number = {3},
	urldate = {2026-07-04},
	journal = {Journal of the Royal Statistical Society Series B: Statistical Methodology},
	author = {Del Moral, Pierre and Doucet, Arnaud and Jasra, Ajay},
	month = jun,
	year = {2006},
	pages = {411--436},
}

@article{hill_1973_DiversityEvennessUnifying,
	title = {Diversity and {Evenness}: {A} {Unifying} {Notation} and {Its} {Consequences}},
	volume = {54},
	copyright = {http://onlinelibrary.wiley.com/termsAndConditions\#vor},
	issn = {0012-9658, 1939-9170},
	shorttitle = {Diversity and {Evenness}},
	doi = {10.2307/1934352},
	language = {en},
	number = {2},
	urldate = {2026-07-04},
	journal = {Ecology},
	author = {Hill, M. O.},
	month = mar,
	year = {1973},
	pages = {427--432},
}

@article{pedregosa_2011_ScikitlearnMachineLearning,
	title = {Scikit-learn: {Machine} {Learning} in {Python}},
	volume = {12},
	url = {http://jmlr.org/papers/v12/pedregosa11a.html},
	number = {85},
	journal = {Journal of Machine Learning Research},
	author = {Pedregosa, Fabian and Varoquaux, Gael and Gramfort, Alexandre and Michel, Vincent and Thirion, Bertrand and Grisel, Olivier and Blondel, Mathieu and Prettenhofer, Peter and Weiss, Ron and Dubourg, Vincent and Vanderplas, Jake and Passos, Alexandre and Cournapeau, David and Brucher, Matthieu and Perrot, Matthieu and Duchesnay, Edouard},
	year = {2011},
	pages = {2825--2830},
}

@article{fisher_1953_DispersionSphere,
	title = {Dispersion on a {Sphere}},
	volume = {217},
	issn = {1364-5021, 1471-2946},
	doi = {10.1098/rspa.1953.0064},
	language = {en},
	number = {1130},
	urldate = {2022-08-16},
	journal = {Proceedings of the Royal Society A: Mathematical, Physical and Engineering Sciences},
	author = {Fisher, R.},
	month = may,
	year = {1953},
	pages = {295--305},
}

@book{rosenberg_1997_LaplacianRiemannianManifold,
	edition = {1},
	title = {The {Laplacian} on a {Riemannian} {Manifold}: {An} {Introduction} to {Analysis} on {Manifolds}},
	isbn = {978-0-521-46300-3 978-0-521-46831-2 978-0-511-62378-3},
	shorttitle = {The {Laplacian} on a {Riemannian} {Manifold}},
	doi = {10.1017/CBO9780511623783},
	urldate = {2026-06-26},
	publisher = {Cambridge University Press},
	author = {Rosenberg, Steven},
	month = jan,
	year = {1997},
}

@article{martino_2017_EffectiveSampleSize,
	title = {Effective sample size for importance sampling based on discrepancy measures},
	volume = {131},
	issn = {01651684},
	doi = {10.1016/j.sigpro.2016.08.025},
	language = {en},
	urldate = {2026-06-26},
	journal = {Signal Processing},
	author = {Martino, Luca and Elvira, Víctor and Louzada, Francisco},
	month = feb,
	year = {2017},
	pages = {386--401},
}

@article{leinster_2012_MeasuringDiversityImportance,
	title = {Measuring diversity: the importance of species similarity},
	volume = {93},
	issn = {0012-9658, 1939-9170},
	shorttitle = {Measuring diversity},
	doi = {10.1890/10-2402.1},
	language = {en},
	number = {3},
	urldate = {2026-06-26},
	journal = {Ecology},
	author = {Leinster, Tom and Cobbold, Christina A.},
	month = mar,
	year = {2012},
	pages = {477--489},
}

@article{lebrigant_2019_ApproximationDensitiesRiemannian,
	title = {Approximation of {Densities} on {Riemannian} {Manifolds}},
	volume = {21},
	issn = {1099-4300},
	doi = {10.3390/e21010043},
	language = {en},
	number = {1},
	urldate = {2026-06-26},
	journal = {Entropy},
	author = {Le Brigant, Alice and Puechmorel, Stéphane},
	month = jan,
	year = {2019},
	pages = {43},
}

@article{pelletier_2005_KernelDensityEstimation,
	title = {Kernel density estimation on {Riemannian} manifolds},
	volume = {73},
	issn = {01677152},
	doi = {10.1016/j.spl.2005.04.004},
	language = {en},
	number = {3},
	urldate = {2026-06-26},
	journal = {Statistics \& Probability Letters},
	author = {Pelletier, Bruno},
	month = jul,
	year = {2005},
	pages = {297--304},
}

@article{li_2011_KNearestNeighborBased,
	title = {k-{Nearest} {Neighbor} {Based} {Consistent} {Entropy} {Estimation} for {Hyperspherical} {Distributions}},
	volume = {13},
	issn = {1099-4300},
	doi = {10.3390/e13030650},
	language = {en},
	number = {3},
	urldate = {2026-06-26},
	journal = {Entropy},
	author = {Li, Shengqiao and Mnatsakanov, Robert M. and Andrew, Michael E.},
	month = mar,
	year = {2011},
	pages = {650--667},
}

@article{kent_1982_FisherBinghamDistributionSphere,
	title = {The {Fisher}-{Bingham} {Distribution} on the {Sphere}},
	volume = {44},
	issn = {1369-7412, 1467-9868},
	doi = {10.1111/j.2517-6161.1982.tb01189.x},
	language = {en},
	number = {1},
	urldate = {2026-06-26},
	journal = {Journal of the Royal Statistical Society Series B: Statistical Methodology},
	author = {Kent, John T.},
	month = sep,
	year = {1982},
	pages = {71--80},
}

@article{hall_1987_KernelDensityEstimation,
	title = {Kernel density estimation with spherical data},
	volume = {74},
	issn = {0006-3444, 1464-3510},
	doi = {10.1093/biomet/74.4.751},
	language = {en},
	number = {4},
	urldate = {2026-06-26},
	journal = {Biometrika},
	author = {Hall, Peter and Watson, G. S. and Cabrera, Javier},
	year = {1987},
	pages = {751--762},
}

@book{grigoryan_2009_HeatKernelAnalysis,
	address = {Providence, R.I},
	series = {{AMS}/{IP} studies in advanced mathematics},
	title = {Heat {Kernel} and {Analysis} on {Manifolds}},
	isbn = {978-0-8218-4935-4},
	language = {eng},
	number = {v. 47},
	publisher = {American Mathematical Society},
	author = {Grigoryan, Alexander},
	year = {2009},
}

@misc{garcia-portugues_2018_OverviewUniformityTests,
	title = {An overview of uniformity tests on the hypersphere},
	copyright = {Creative Commons Attribution Non Commercial Share Alike 4.0 International},
	url = {https://arxiv.org/abs/1804.00286},
	doi = {10.48550/ARXIV.1804.00286},
	urldate = {2026-06-26},
	publisher = {arXiv},
	author = {García-Portugués, Eduardo and Verdebout, Thomas},
	year = {2018},
}

@article{banerjee_2005_ClusteringUnitHypersphere,
	title = {Clustering on the unit hypersphere using von mises-fisher distributions},
	volume = {6},
	url = {http://jmlr.org/papers/v6/banerjee05a.html},
	number = {46},
	journal = {Journal of Machine Learning Research},
	author = {Banerjee, Arindam and Dhillon, Inderjit S. and Ghosh, Joydeep and Sra, Suvrit},
	year = {2005},
	pages = {1345--1382},
}

@article{muandet_2017_KernelMeanEmbedding,
	title = {Kernel {Mean} {Embedding} of {Distributions}: {A} {Review} and {Beyond}},
	volume = {10},
	issn = {1935-8237, 1935-8245},
	shorttitle = {Kernel {Mean} {Embedding} of {Distributions}},
	url = {http://www.nowpublishers.com/article/Details/MAL-060},
	doi = {10.1561/2200000060},
	language = {en},
	number = {1-2},
	urldate = {2021-11-18},
	journal = {Foundations and Trends® in Machine Learning},
	author = {Muandet, Krikamol and Fukumizu, Kenji and Sriperumbudur, Bharath and Schölkopf, Bernhard},
	year = {2017},
	pages = {1--141},
}

@article{gretton_2012_KernelTwoSampleTest,
	title = {A {Kernel} {Two}-{Sample} {Test}},
	volume = {13},
	issn = {1532-4435},
	number = {null},
	journal = {Journal of Machine Learning Research},
	publisher = {JMLR.org},
	author = {Gretton, Arthur and Borgwardt, Karsten M. and Rasch, Malte J. and Schölkopf, Bernhard and Smola, Alexander},
	month = mar,
	year = {2012},
	pages = {723--773},
}

@book{mardia_2000_DirectionalStatistics,
	address = {Chichester; New York},
	series = {Wiley {Series} in {Probability} and {Statistics}},
	title = {Directional statistics},
	isbn = {978-0-471-95333-3},
	publisher = {J. Wiley},
	author = {Mardia, K. V. and Jupp, Peter E.},
	year = {2000},
}

@book{chung_1997_SpectralGraphTheory,
	address = {Providence, R.I},
	series = {Regional conference series in mathematics},
	title = {Spectral graph theory},
	isbn = {978-0-8218-0315-8},
	number = {no. 92},
	publisher = {Published for the Conference Board of the mathematical sciences by the American Mathematical Society},
	author = {Chung, Fan R. K.},
	year = {1997},
	note = {chung\_1997\_SpectralGraphTheory},
}

@article{coifman_2006_DiffusionMaps,
	title = {Diffusion maps},
	volume = {21},
	issn = {10635203},
	url = {https://linkinghub.elsevier.com/retrieve/pii/S1063520306000546},
	doi = {10.1016/j.acha.2006.04.006},
	language = {en},
	number = {1},
	urldate = {2021-11-17},
	journal = {Applied and Computational Harmonic Analysis},
	author = {Coifman, Ronald R. and Lafon, Stéphane},
	month = jul,
	year = {2006},
	note = {coifman\_2006\_DiffusionMaps},
	pages = {5--30},
}
